\pdfoutput=1

\documentclass[11pt]{article}

\usepackage[]{ACL}
\usepackage{float}

\usepackage{placeins}

\usepackage{times}
\usepackage{latexsym}
\usepackage{graphicx}
\usepackage{placeins} 

\usepackage[T1]{fontenc}

\usepackage[utf8]{inputenc}

\usepackage{microtype}

\usepackage{inconsolata}

\usepackage{amsmath}
\usepackage{amssymb}
\usepackage[dvipsnames]{xcolor}
\usepackage[ruled,linesnumbered]{algorithm2e}
\usepackage{enumitem}
\usepackage{svg}
\usepackage{adjustbox}
\usepackage{booktabs}
\usepackage{caption}
\usepackage{titlesec}

\titlespacing\paragraph{0pt}{0.5ex}{1.3ex plus 0.1ex}

\usepackage{array}
\usepackage{graphicx}
\usepackage{stfloats}
\usepackage{longtable}
\usepackage{afterpage}

\title{Reasoning Models Reason Well, Until They Don't}

\author{
Revanth Rameshkumar$^{1}$,
Jimson Huang$^{2}$,
Yunxin Sun$^{2}$,
Fei Xia$^{1}$,
Abulhair Saparov$^{2}$\\
$^{1}$University of Washington\\
$^{2}$Purdue University\\
\texttt{\{revr,fxia\}@uw.edu \{huan2073,sun1114,asaparov\}@purdue.edu}
}

\begin{document}
\maketitle
\begin{abstract}
    Large language models (LLMs) have shown significant progress in reasoning tasks. However, recent studies show that transformers and LLMs fail catastrophically once reasoning problems exceed modest complexity. We revisit these findings through the lens of large reasoning models (LRMs)---LLMs fine-tuned with incentives for step‑by‑step argumentation and self‑verification. LRM performance on graph and reasoning benchmarks such as \texttt{NLGraph} seem extraordinary, with some even claiming they are capable of generalized reasoning and innovation in reasoning-intensive fields such as mathematics, physics, medicine, and law. However, by more carefully scaling the complexity of reasoning problems, we show existing benchmarks actually have limited complexity. We develop a new dataset, the \texttt{Deep Reasoning Dataset} (\texttt{DeepRD}), along with a generative process for producing unlimited examples of scalable complexity. We use this dataset to evaluate model performance on graph connectivity and natural language proof planning. We find that the performance of LRMs drop abruptly at sufficient complexity and do not generalize. We also relate our LRM results to the distributions of the complexities of large, real-world knowledge graphs, interaction graphs, and proof datasets. We find the majority of real-world examples fall inside the LRMs' success regime, yet the long tails expose substantial failure potential. Our analysis highlights the near-term utility of LRMs while underscoring the need for new methods that generalize beyond the complexity of examples in the training distribution.
\end{abstract}

\section{Introduction}
Large language models (LLMs) have recently demonstrated impressive performance on reasoning-intensive tasks. This is, in a large part, due to methods such as reinforcement learning with verifiable rewards (RLVR)  \cite{lambert2025tulu3pushingfrontiers}, thereby producing large reasoning models (LRMs).
LRMs seemingly excel at coding, mathematics, and small graph puzzles \cite{openai2025introducing-o3, deepseekai2025deepseekr1incentivizingreasoningcapability}.
Some proponents claim that the reasoning abilities of LRMs have advanced to the point of being capable of performing original scientific research \cite{jansen2025codescientistendtoendsemiautomatedscientific}, \cite{zheng2025deepresearcherscalingdeepresearch, xu2025comprehensivesurveydeepresearch}.
However, despite improved performance on moderately more complex problems, it is less clear whether LRMs can generalize to \emph{highly-complex} reasoning problems.
Scaling experiments on transformers suggest that they have difficulty learning tasks such as multi-hop search, graph coloring, or finding Hamiltonian paths \citep{saparov2024transformersstrugglelearnsearch, wang2024languagemodelssolvegraph, heyman2025evaluatingsystematicreasoningabilities}---suggesting fundamental limitations for LLMs and LRMs.
If LRMs cannot generalize to more complex reasoning tasks, then it is unlikely that they will succeed in more reasoning-intensive applications such as automated research. Thus, establishing a performance baseline on complex reasoning is important, but not without challenges: existing corpora and benchmarks (i) might be present in the models' training data and (ii) complexity of existing benchmarks is limited.

Real-world reasoning tasks, including those in natural language, do not have a fundamental upper bound in their complexity.
There are many real-world domains where completing queries or tasks require deeper reasoning, such as in multi-hop question-answering \cite{saxena-etal-2020-improving} where many reasoning steps are required to answer questions correctly. For example ``How many asteroids were found between one of Jupiter's large Trojan asteroids and 4239 Goodman?'' would require 177 reasoning steps (or ``hops'') in \texttt{Wikidata5M} \cite{wang-etal-2021-kepler}. Relatedly, biomedical queries such as ``How are the genes of tuberculosis bacteria used in lung cancer research?'' \cite{Xu_2025}, or about drug interaction that can involve reasoning over complex drug interaction graphs \cite{Wishart2018DrugBank}.
In instruction following, narrative understanding, and dialogue planning, a reasoning system might need to consider very long chains of events to properly characterize or summarize events. For example, in \citet{rameshkumar-bailey-2020-storytelling}, there are dialogue chains that are over 2000 turns long for a turn-based roleplaying game---a model might need make long traversals to piece together narrative information. To answer questions like ``What motivated a character to pick up the sword?'', a traversal of the character's actions may be required. 
Natural language proof generation, producing strictly connected logical inference chains, and program synthesis are also problems with potentially unbounded complexity \cite{welleck2022naturalprovergroundedmathematicalproof,chen2023teachinglargelanguagemodels}. 

\begin{figure*}[ht]
     \vspace{-0.1em}
     \hspace{-1.2em}\adjustbox{trim=3em 10em 3em 5.7em}{
         \includegraphics[width=1.15\textwidth]{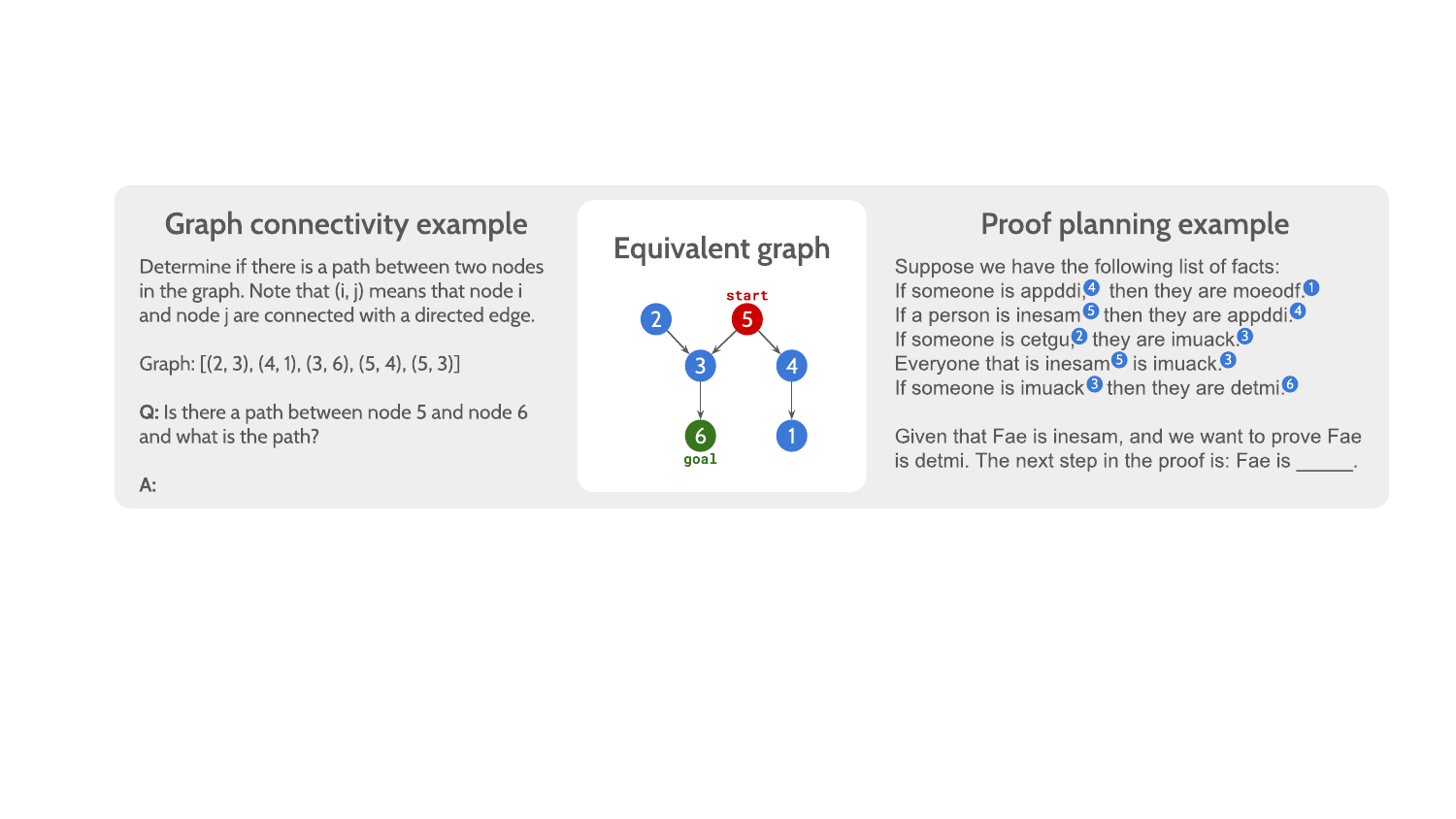}
     }
     \vspace{-0.5em}
     \caption{An example from \texttt{DeepRD}. The example has complexity metrics: lookahead $L=2$ and number of branches $B=2$ (see Section~\ref{sec:complexity} for definitions of complexity metrics). The same underlying graph is transformed into a symbolic graph query for an LLM to find a valid path, and into a natural language query for an LLM to plan the next step in a proof.}
     \label{fig:overview_figure}
     \vspace{-1em}
 \end{figure*}

Recently, \citet{shojaee2025illusionthinkingunderstandingstrengths} evaluated LRMs on arbitrarily complex reasoning problems by using four puzzle scenarios. However, an exponential number of steps is required to solve some of the puzzles, and so the inability to solve the puzzles may be due to model token limits rather than reasoning limitations \cite{lawsen2025commentillusionthinkingunderstanding}. Although the puzzles were complex in terms of the number of steps required to solve them, they were simple to describe, and LRMs are able to generate code to programmatically solve the puzzles. Therefore, more analysis is needed on more realistic reasoning problems. Particularly for scenarios like natural language deduction and proof planning which is not as easily solved programmatically.

To test the limits of LRM reasoning, we query models to solve synthetically-generated graph reasoning and proof planning tasks with unbounded, but controllable, complexity. This new dataset, called \texttt{Deep Reasoning Dataset} or \texttt{DeepRD}, contains problems with widely varying complexity (in terms of the number of required reasoning steps; hence ``deep''), but are simultaneously simple to generate synthetically. Thus, the examples serve as a ``lower bound'' on the complexity of realistic reasoning problems, and a reasoning system would need to perform well on \texttt{DeepRD} if they are able to solve similarly complex and novel real-world reasoning tasks.

\paragraph{Research Questions and Contributions}
In this paper, we study the following research questions:
\begin{itemize}[noitemsep,topsep=3pt,leftmargin=1.2em]
\item \textbf{RQ1} How does LLM/LRM performance on graph reasoning scale with complexity?
\item \textbf{RQ2} How does performance scale on natural language proof planning?
\item \textbf{RQ3} How complex are reasoning problems found ``in the wild'' or in the training data of LLMs and LRMs?
\end{itemize}
We make the following contributions\footnote{All code, data, and model responses are open source at \url{https://github.com/RevanthRameshkumar/DeepRD}.}:
\begin{itemize}[noitemsep,topsep=3pt,leftmargin=1.2em]
\item \textbf{C1} We comprehensively characterize the performance of LLMs and LRMs on symbolic graph connectivity problems with well-defined complexity metrics.
\item \textbf{C2} We evaluate model performance when the same graph structures in RQ1 and C1 are encoded as natural language axioms, in order to mirror a proof planning task.
\item \textbf{C3} We introduce the \texttt{\textbf{Deep Reasoning Dataset}} (\texttt{DeepRD}), which contains 2220 examples of symbolic graph connectivity and proof planning examples; along with the parameterized generator for producing unlimited synthetic examples of any requested complexity.
\item \textbf{C4} We carry out a detailed analysis of the complexity distribution of a variety of real-world reasoning scenarios (graphs and proofs), with results suggesting that LRMs are not able to generalize beyond the complexity of examples that typically occur in their training.
\item \textbf{C5} We evaluate LRMs on real-world natural language proofs from the \texttt{NaturalProofs} dataset, showing that their ability to detect errors and validate proofs exhibits the same depth-sensitive performance patterns as on the synthetic \texttt{DeepRD} dataset, where the accuracy sharply drops as the proof length increases.

\end{itemize}

\section{Related Work}
 
\paragraph{Evaluation of LLMs on Graph Reasoning}
Recent work has demonstrated that both standard transformers and large language models (LLMs) face significant challenges when applied to graph reasoning tasks. For example, recent studies show that even state-of-the-art LLMs struggle with explicit graph reasoning in various tasks as the complexity of the graph increases modestly \cite{wang2024languagemodelssolvegraph,agrawal2024llmsperformstructuredgraph,zhang2024llm4dyglargelanguagemodels}. This observation is further supported by \citet{saparov2024transformersstrugglelearnsearch}, who found that transformers have difficulty learning robust search strategies over large, complex graphs---likely due to architectural limitations, and that performance does not scale well with model size. Similarly, \citet{fatemi2023talklikegraphencoding} highlights that performance on graph reasoning tasks is highly sensitive to the chosen text encoding of the graph, as well as the inherent complexity of the graph task itself. While LRMs seem to be better at graph reasoning than baseline LLMs, they still fail at reasoning on high-complexity examples \cite{heyman2025evaluatingsystematicreasoningabilities}. We show that these high complexities are found in real-world knowledge graph and graph datasets.

\paragraph{Prompting and Post-training for Reasoning}  
Explicit reasoning techniques like chain‑of‑thought guide LLMs through intermediate steps, reducing hallucinations and boosting performance on complex tasks \cite{kim2023cotcollectionimprovingzeroshot,jin2024graphchainofthoughtaugmentinglarge}. However, even with self‑consistency and verification strategies \cite{wang2023selfconsistencyimproveschainthought,wang2024languagemodelssolvegraph} or the selection‑inference framework \cite{creswell2022selectioninferenceexploitinglargelanguage}, large models still fail to generalize on high‑complexity reasoning tasks \cite{saparov2024transformersstrugglelearnsearch}. Reinforcement learning approaches---exemplified by OpenAI’s \texttt{o3} \cite{openai2025introducing-o3} and \texttt{DeepSeek‑R1} \cite{deepseekai2025deepseekr1incentivizingreasoningcapability}---use verifiable rewards to incentivize step‑by‑step reasoning, outperforming fine-tuned models yet still erring on moderate‑complexity reasoning tasks \cite{heyman2025evaluatingsystematicreasoningabilities}.

\paragraph{Proof Planning with LLMs}
LLMs excel at individual deductions but collapse on complete proof planning once multiple potential deduction steps are available \cite{saparov2023languagemodelsgreedyreasoners}. Grounded generators like \texttt{NaturalProver} improve short (2–6 step) proofs via retrieval and constrained decoding \cite{welleck2022naturalprovergroundedmathematicalproof}. Structure-aware demonstration with pruning can delay (but not prevent) errors at increased complexity \cite{zheng2025exploringrolereasoningstructures}. Inference‐time methods such as \texttt{LogicTree} use caching and linearized premise selection to maintain coherence \cite{he2025logictreestructuredproofexploration} and multi‑agent frameworks like \texttt{MA‑LoT} separate high‑level natural language planning from formal verification, yet still exhibit steep performance drops as proofs deepen \cite{wang2025malotmodelcollaborationleanbasedlong}. These observations necessitate a more systematic evaluation of LLM and LRM reasoning ability with respect to proof planning complexity.

\paragraph{Generalization in Reasoning}
Prior work has probed generalization in reasoning from complementary angles—systematic rule recombination in short narratives \cite{sinha-etal-2019-clutrr}, planning in NP-hard settings and LRM–LLM gaps \cite{valmeekam2024llmscantplanlrms}, qualitative spatial/temporal composition with disjunctive paths \cite{khalid-etal-2025-large}, and teleological accounts of next-token prediction \cite{mccoy2023embersautoregressionunderstandinglarge}. Our contribution is a structural analysis that controls for complexity via parameterization of graph depth and fan-out, provides a contamination-free evaluation of reasoning over directed acyclic graphs (DAGs) which were created using our parameterized framework, shows sharp depth-correlated failures even when the graph is a chain (no branching), and enables future exploration through the ability to control the complexity of graph and proof reasoning queries. We further manually inspect the full reasoning traces of LLMs and LRMs, and comprehensively categorize their errors into several error types.

\section{Methodology}
In this section, we describe the reasoning tasks on which we evaluate LLMs and LRMs. We also define metrics to measure the complexity of reasoning problems, which we utilize in our study of LLM/LRM reasoning ability as a function of example complexity.

\subsection{Reasoning Tasks} \label{sec:reasoning_tasks}

There are many reasoning tasks that can be used for evaluation. Since we endeavor to measure model reasoning ability vs complexity, we need a task for which examples are easily generated synthetically, a well-defined metric of complexity is available, and each example is not intractably difficult (i.e., where the number of required steps is polynomial in the complexity). Graph connectivity is a task where the goal is to find a path between two nodes in a graph. We choose to use the graph connectivity task since it is intuitive, it can be solved in linear time with respect to the problem complexity, and the problem complexity is readily controllable. This also allows us to controllably test for generalization at different example complexities while maintaining a reasonable context size (in order to avoid accuracy drops due to context limits rather than errors in reasoning; \citealt{modarressi2025nolimalongcontextevaluationliteral}).

We also use a simple proof planning task in deductive reasoning. Here, each example contains a list of facts of the form ``if $A$ then $B$'' (e.g., ``If someone is a cat, then they are a mammal''). Each example also contains a start fact and a goal fact (e.g., ``Given Gwendolyn is a cat, we want to prove that Gwendolyn is warm-blooded''). The model is then queried to give the next step in the proof. Examples of this task are equivalent to examples of graph connectivity, where each node in a graph can be mapped to a clause (e.g., $v_1$ is mapped to ``someone is a cat''), and each edge is mapped to a fact (e.g., the edge $v_1 \to v_2$ is mapped to ``If someone is a cat, then they are a mammal''). Thus, we are able to generate proof planning examples by first generating a graph connectivity example, and then mapping each into natural language (see Section~\ref{appendix:translation} for details). Since these examples are expressed in unstructured natural language, they are not easily solved programmatically, and similarly, it is more difficult to automatically evaluate the model output. Due to this, in addition to the fact that proof planning examples are substantially longer, we simplify this task: Rather than asking the model to generate the full proof, we only require it to produce the next proof step.

\subsection{Reasoning Task Complexity} \label{sec:complexity}

In order to test models on examples of higher complexity, we require a well-defined measure of the complexity of reasoning examples. To do so, we consider the graph analogy of the proof planning task: In order to find the next step along the path from the start node to the goal node, a model must search for paths between the two nodes.
A common and intuitive way to measure the complexity of a graph search task is by the distance from the start node to the goal node.
A simple breadth-first search (BFS; Algorithm~\ref{alg:bfs} in Section~\ref{app:lookahead}) can be used to compute the distance from the start node to the goal node.
However, a sufficiently powerful model could learn a more efficient algorithm to find the next node along the path from the start to the goal node. Consider a chain graph (a graph of nodes connected one after the other) in which the distance from the start to the goal node is $d$. A more clever model could surmise that since the start node has only one child, there is only one possible next node toward the goal. Such a model would find the correct next step in this search problem trivially (termed the ``Clever Hans cheat'' in \citealt{DBLP:conf/icml/BachmannN24}). Therefore, we use the \textbf{lookahead} metric $L$ from \citet{saparov2024transformersstrugglelearnsearch} to measure the complexity of this task, which is defined as the number of BFS iterations required to determine the next correct node along the path from the start node to the goal node, where the search stops early whenever it determines that any path to the goal must pass through one of the children of the source node. Algorithm~\ref{alg:lookahead} in Section~\ref{app:lookahead} demonstrates how to use BFS to compute the lookahead.
In other words, for chain graphs the lookahead is simply 1 since the model only needs one BFS iteration to determine that there is only one correct answer. In Figure~\ref{fig:overview_figure}, the model must perform 2 BFS iterations from the start node to determine whether to proceed to node 3 or to node 4.
Closely related to lookahead, the \textbf{number of branches} $B$ is defined as the number of outgoing nodes from the start node. Examples with larger $B$ may be more difficult as there are a larger number of possible next steps and random guessing is less effective.

\section{Evaluating Models on Highly Complex Reasoning}
In this section, we determine how LLMs and LRMs perform on aforementioned reasoning tasks with increasing complexity.

\subsection{Models and Data}

\paragraph{Models} LRMs are trained via reinforcement learning with verifiable rewards \cite{deepseekai2025deepseekr1incentivizingreasoningcapability}. We evaluate a handful of LLMs and LRMs: (1) DeepSeek's \texttt{R1} (LRM), (2) its base model \texttt{V3} (LLM), (3) OpenAI's \texttt{o3-mini} (LRM), and (4) \texttt{GPT-4o} (LLM).
We also evaluate the full version of \texttt{o3} on small subsets of \texttt{DeepRD}, due to budget constraints. See Section~\ref{appendix_modelsettings} for all model settings.

\paragraph{\texttt{NLGraph}}
The first dataset we evaluate on is \texttt{NLGraph} \citep{wang2024languagemodelssolvegraph}, which is a benchmark designed to test graph reasoning in LLMs across many tasks, including graph connectivity. The examples in \texttt{NLGraph} were generated using the Erd\H{o}s--R\'enyi distribution \citep{Erdos1959}. Each of the resulting graphs were transformed into an LLM input with the graph expressed as a list of edges. The test cases are split into \texttt{easy}, \texttt{medium}, and \texttt{hard} categories in terms of difficulty (by node count). In the \texttt{hard} category, there are 7770 graphs.

We emphasize that despite the fact that graphs in \texttt{NLGraph-hard} have a large number of nodes, the examples are fairly simple in terms of lookahead.
Since the lookahead of any example is bounded above by path length, the expected lookahead is less than or equal to the expected path length. The average path length in Erd\H{o}s--R\'enyi graphs is $\frac{\log N - \gamma}{\log(pN)} + \frac{1}{2}$ where $N$ is the number of nodes, $p$ is the edge probability, and $\gamma$ is the Euler-Mascheroni constant (Eq.~16 in \citealt{Fronczak_2004}). Since $p\ge 0.3$ and $N \ge 26$ in \texttt{NLGraph-hard}, the expected lookahead of \texttt{NLGraph-hard} examples is bounded by $\sim 1.805$. We only evaluate the reasoning models on the \texttt{hard} problems for brevity.

\vspace{-0.2em}
\paragraph{\texttt{Deep Reasoning Dataset} (\texttt{DeepRD})} \citet{saparov2024transformersstrugglelearnsearch} shows that transformers fail to learn to search on graphs with large lookaheads, and model scale does not alleviate this limitation. In order to test this claim for LLMs and LRMs, we modify the data generation process from \citet{saparov2024transformersstrugglelearnsearch} to generate graph connectivity examples with many lookaheads (ranging from 2 through 800). \citet{saparov2024transformersstrugglelearnsearch} does not aim to generate graphs that approach the same size as ours, even though we use their formulations of complexity as the foundation of our evaluation method. We modify the original generation process to more efficiently generate very large graphs and to reduce the likelihood of generating isomorphisms due to star patterns (graph with single central node with independent branches). We also provide a framework around this algorithm to enable a tightly controlled, depth-parameterized testbed.
For each lookahead, and for each branch size $B\in\{2,4,8,16\}$, we generate an example.
We take care not to generate isomorphic graphs by utilizing the \texttt{NetworkX} function \href{https://networkx.org/documentation/stable/reference/algorithms/generated/networkx.algorithms.isomorphism.faster_could_be_isomorphic.html}{\texttt{faster\_could\_be\_isomorphic}} \cite{hagberg2008exploring}.
We sample until every branch-lookahead pair has 10 examples. We also generate chain graph examples---where each graph is linear, consisting of a single path (i.e., branch size $B=1$)---with a variety of depths (ranging from 2 through 1536) for further analysis.\footnote{We use ``depth'' rather than lookahead to describe the length of chain examples, since by definition, the lookahead of any chain graph is 1.} In these chain examples, the only step required of the model should be to find the next connected node (no search or backtracking is required). Because of these steps, our dataset is also designed to detect when the model relies heavily on search heuristics. We believe a model which is overly reliant on search heuristics to the detriment of accuracy should be considered lacking in reasoning ability. Once the examples are sampled from the generator, we convert them into textual prompts, as in \texttt{NLGraph} (Figure \ref{fig:overview_figure}). We also convert each example into a proof planning example, as described in Section~\ref{sec:reasoning_tasks}.
In total we generate 2220 examples for evaluation. For future experiments, the same data generation process can be used to generate indefinitely more graphs. We release this evaluation data, and the generation process, as the \texttt{Deep Reasoning Dataset} (\texttt{DeepRD}).

\subsection{Reasoning about Graph Connectivity}

\begin{table}[t]
\centering
\small
\setlength{\tabcolsep}{2pt}

  \begin{minipage}[t]{0.45\linewidth}
    \vspace{0pt}
    \centering
    \renewcommand{\arraystretch}{0.85}
    \begin{tabular}{lc}
      \toprule
      \multicolumn{2}{l}{\textbf{Original} (\texttt{text-davinci-003})} \\
      \midrule
      Prompting    & Accuracy \\ 
      \midrule
      Zero-shot & 63\% \\
      Few-shot  & 77\% \\
      CoT       & 77\% \\
      0-CoT     & 69\% \\
      CoT+SC    & 83\% \\
      \bottomrule
    \end{tabular}
  \end{minipage}
  \hfill
  \begin{minipage}[t]{0.45\linewidth}
    \vspace{0pt}
    \centering
    \renewcommand{\arraystretch}{0.85}
    \begin{tabular}{lc}
      \toprule
      \multicolumn{2}{l}{\textbf{New (Multiple Models)}} \\
      \midrule
      Model        & Accuracy \\ 
      \midrule
      \texttt{GPT-4o}       & 75\% \\
      \texttt{o3-mini}  & 99\% \\
      DeepSeek \texttt{V3}  & 79\% \\
      DeepSeek \texttt{R1}  & 96\% \\
      \bottomrule
    \end{tabular}
  \end{minipage}

\vspace{-0.6em}
\caption{(\textbf{left}) The \emph{original} average accuracies on \texttt{NLGraph-hard} obtained by \cite{wang2024languagemodelssolvegraph}; using \texttt{text-davinci-003} as the model where the model is asked to determine whether or not a path exists. (\textbf{right}) The average accuracies of various models on the same data where we ask the model to provide a full path. We find the results of the reasoning models, even with stricter requirements, to be significantly better than \texttt{text-davinci-003} augmented with chain-of-thought (CoT) and self-consistency (SC) prompting.}
\label{tab:nlgraphhard_fullcontext}
\vspace{-1.5em}
\end{table}

\begin{figure}[!t]
    \centering
    \includegraphics[width=\linewidth]{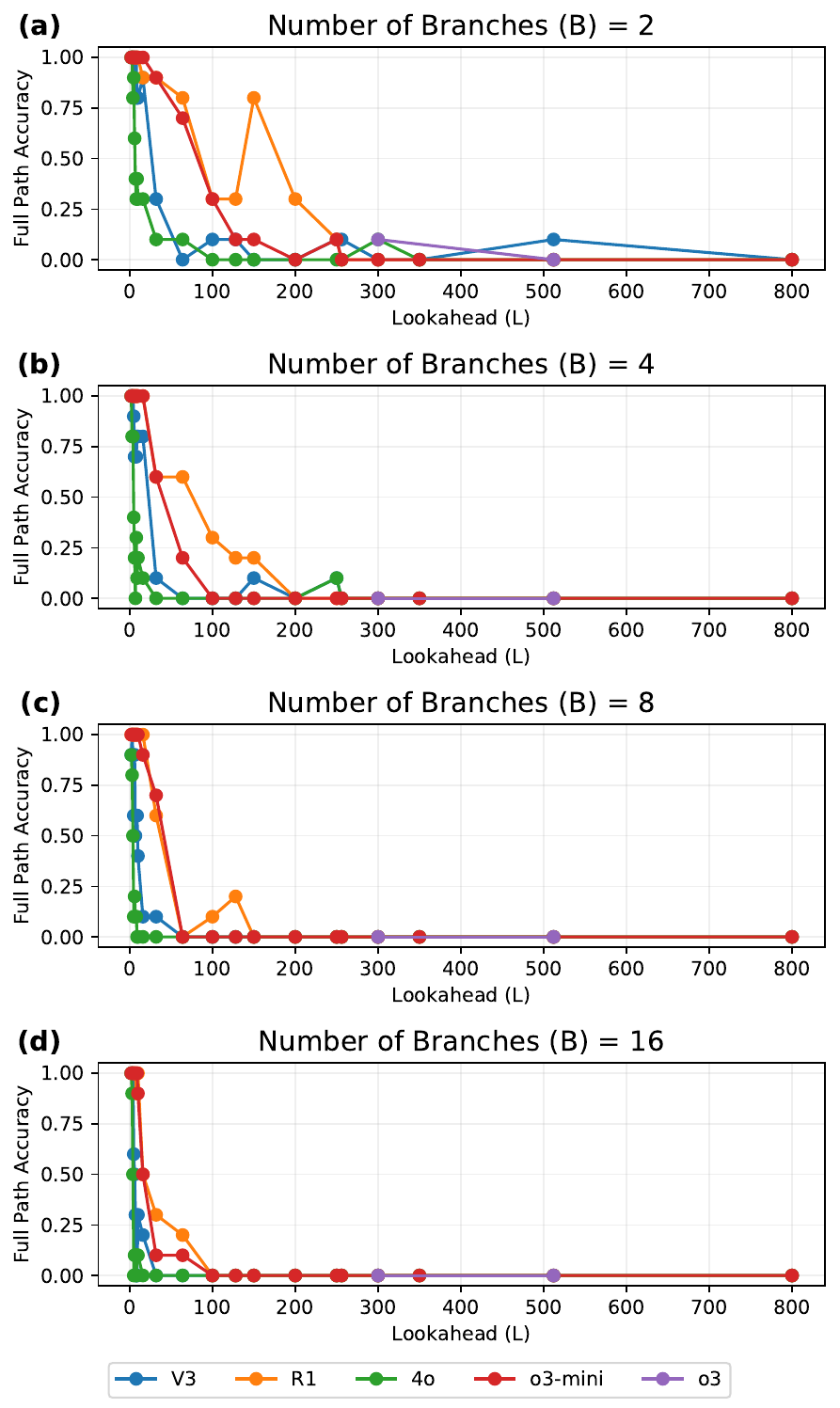}
    \vspace{-1.9em}
    \caption{We evaluate the models \texttt{V3}, \texttt{R1}, \texttt{4o}, \texttt{o3-mini}, and \texttt{o3} on reasoning about graph connectivity. We show the average \emph{full path accuracy} for each $B, L$ pair, for $B= 2, 4, 8, 16$ and $L$ between 2 and 800.}
    \label{fig:lookahead_metrics_4_branches}
    \vspace{-1.8em}
\end{figure}

We establish the baseline performance for graph reasoning as a function of the lookahead complexity measure. First, we show the performance of the LRMs and LLMs on \texttt{NLGraph-hard} (Figure~\ref{tab:nlgraphhard_fullcontext}). We observe that LRMs do significantly better on the connectivity problem compared to LLMs, achieving near-perfect scores. However, as we noted earlier, the lookahead of \texttt{NLGraph} examples are small. So, we want to determine how the performance of LRMs scale on examples with larger lookaheads. In order to do so, we generate graphs using fixed lookaheads of increasingly larger size.

\paragraph{Evaluation}
We use \emph{full path accuracy} as our primary metric.
Critical differences between our evaluation and the one done by \citet{wang2024languagemodelssolvegraph} are that (i) we require a valid path to be produced in addition to the yes/no answer, and that (ii) the goal node is always reachable from the start node.
We use a simple \texttt{GPT-4o-mini} prompt to extract the path from the model's final answer (token after the ``thinking end'' token).
We see in Figure~\ref{fig:lookahead_metrics_4_branches} that all models drop in performance abruptly and collapse to 0 accuracy, with the drops occurring at smaller lookaheads $L$ as the number of branches $B$ increases. We also see that the reasoning models (\texttt{R1} and \texttt{o3-mini}) drop in performance at higher $L$ than the LLMs. We measure the rate at which the models hallucinate non-existent edges (Section~\ref{app:error_analysis}) and find it to be quite high (over 50\% in some cases) but generally trends downwards as $L$ increases. This is due to the model incorrectly concluding that there is no path, or the predicted path is incomplete. LRMs, however, have a much lower edge hallucination rate.
We also ran the full \texttt{o3} model on lookaheads 300 and 512.
Similar to the smaller models, \texttt{o3} gets 10\% accuracy for $B=2$ and otherwise collapses to 0 accuracy.

\begin{figure}[!t]
    \centering
    \includegraphics[width=\linewidth]{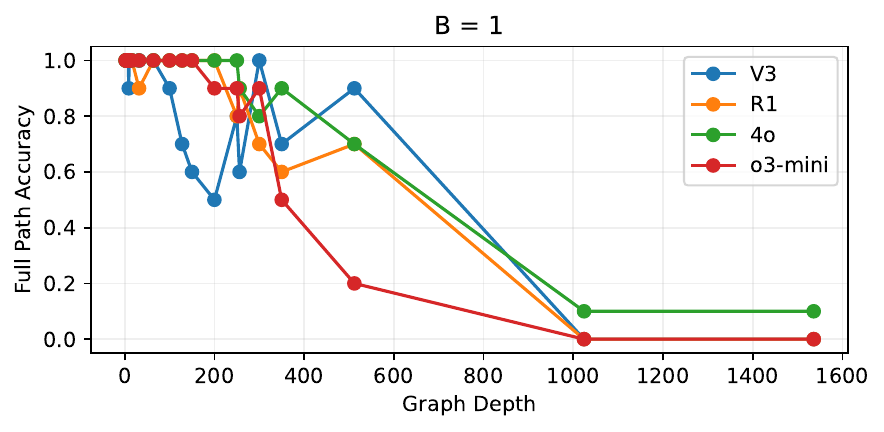}
    \vspace{-2.2em}
    \caption{For the graph connectivity problem, we show the average \emph{full path accuracy} for test cases with graph depths between 2 to 1536, when constraining the graph to be a chain graph (one node after the other).}
    \label{fig:lookahead_metrics_1_branch}
    \vspace{-1.2em}
\end{figure}

\vspace{-0.2em}
\paragraph{Trivial Complexity} We also control for complexity due to search by looking at model performance on graphs with no branches ($B=1$), while increasing graph depth (Figure \ref{fig:lookahead_metrics_1_branch}).
We show that when we simplify the task to this trivial complexity, the models perform significantly better than the $B=2$ case, but still quickly drop in performance for large depths (Figure \ref{fig:lookahead_metrics_1_branch}). Even in this extremely simple reasoning task, the models fail to truly generalize.

\begin{figure}[!t]
    \centering
    \includegraphics[width=\linewidth]{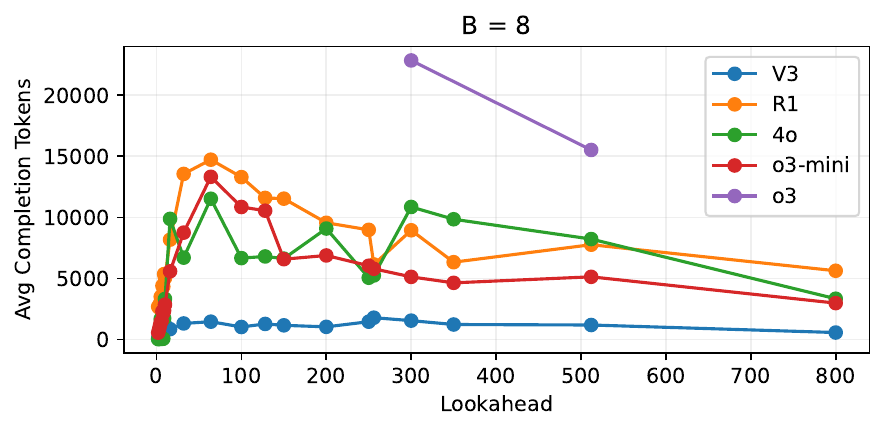}
    \vspace{-2.2em}
    \caption{For the graph connectivity problem, we show the average token completion count for $B=8$ and $L$ between 2 and 800. We can only run the large \texttt{o3} on a small range of $L$ due to budget constraints.}
    \label{fig:token_usage_graph}
    \vspace{-1.3em}
\end{figure}

\paragraph{Token Usage} \citet{lawsen2025commentillusionthinkingunderstanding} claimed that the LRM analysis of \citet{shojaee2025illusionthinkingunderstandingstrengths} was unfair due to token limits, which would limit the model's ability to work through reasoning problems.
Our analysis shows that token limits do not cause the drops in accuracy.
While the LLMs (\texttt{V3}, \texttt{4o}) did hit length limits throughout the set of lookaheads, the LRMs (\texttt{R1}, \texttt{o3-mini}) rarely hit length limits (Section~\ref{app:error_analysis}). We inspected the ``\texttt{stop\_reason}'' field in every model response. In fact, the completion token usage seems to \emph{decrease} with increasing lookahead (Figure~\ref{fig:token_usage_graph}). Results for all branches are shown in Section~\ref{app:additional_analysis}.\footnote{In some cases, the API returns a refusal due to the number of requested tokens being too high. This happens very rarely, except in the $L=800, B=16$ case where the generated examples were very large (full results in Section~\ref{app:error_analysis}). We don't count this as an error, so it does not negatively impact the models' accuracies.}
Manual inspection of the LRM responses for more complex examples also seems to indicate no information is being truncated, as edges from the very beginning of the prompt are present in the output. Interestingly, if the nodes in the inputs for the trivial complexity evaluation were presented in the order of correct traversal, all models achieve near-perfect performance. This further indicates the models are limited by reasoning ability, and not by token limits.

\subsection{Proof Planning in Deductive Reasoning}
Next, we evaluate the reasoning ability of LRMs via proof planning, by rendering each example as a natural language proof planning problem.

\begin{figure}[!t]
    \centering
    \includegraphics[width=\linewidth]{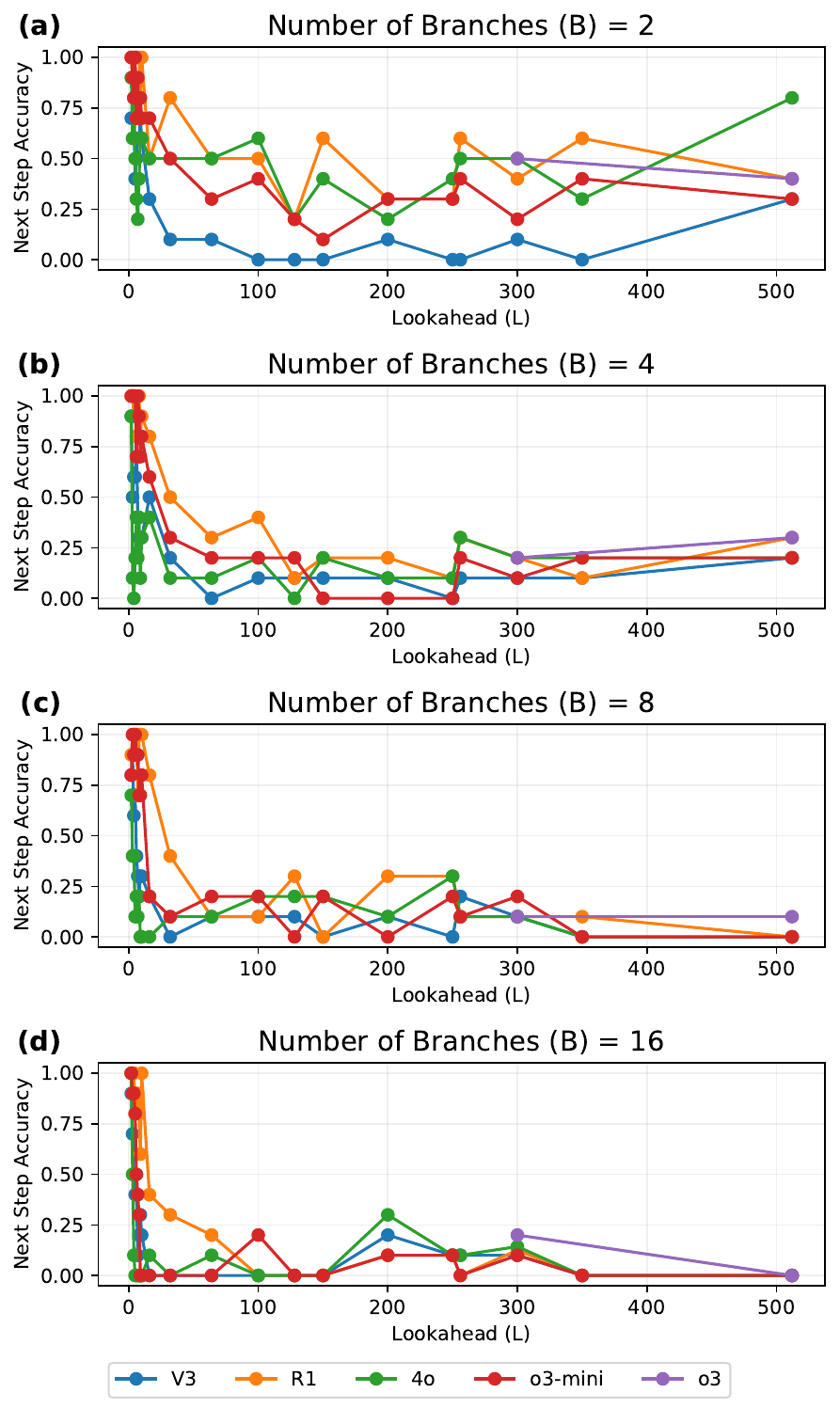}
    \vspace{-1.9em}
    \caption{We evaluate the models on the proof planning task. We show the average \emph{next step prediction accuracy} for each $(B, L)$ pair, for $B= 2, 4, 8, 16$ and $L$ between 2 and 800.}
    \label{natural_language_metrics}
    \vspace{-1.4em}
\end{figure}

\begin{figure*}[!t]
    \centering
    \includegraphics[width=\textwidth]{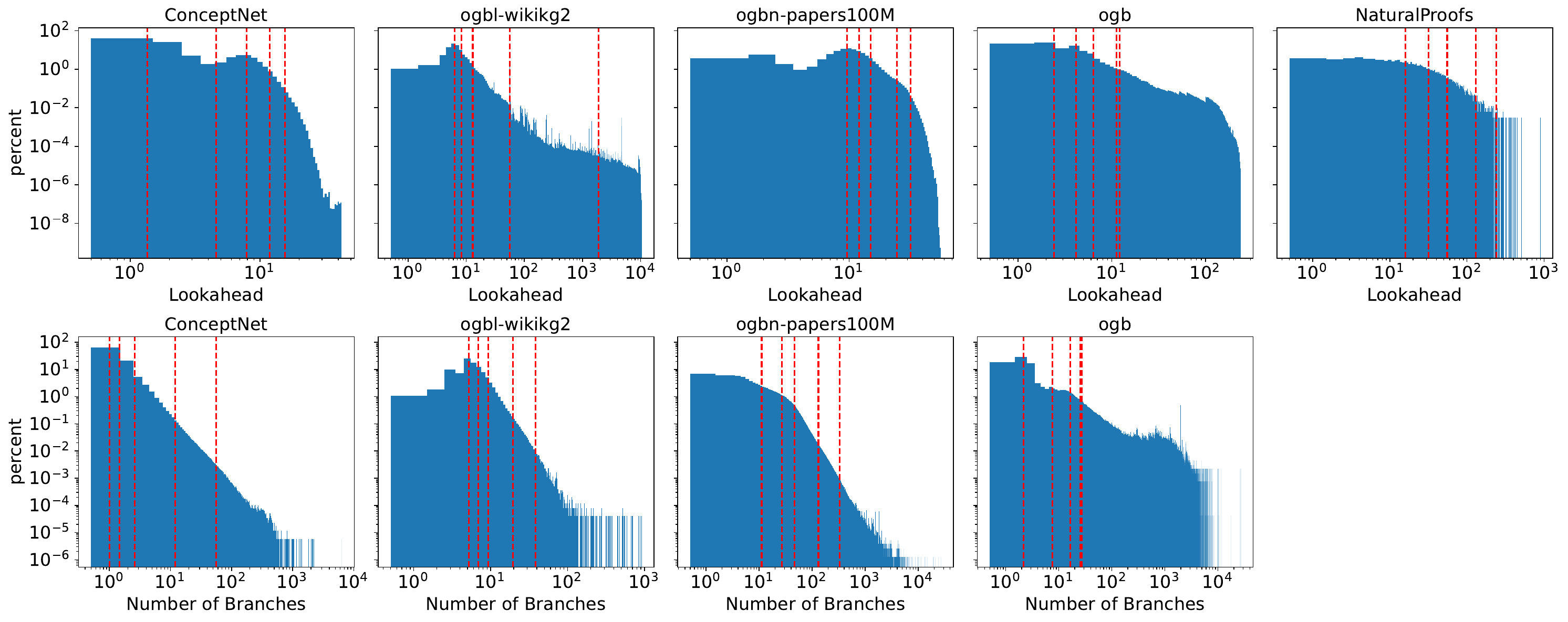}
    \vspace{-1.8em}
    \caption{Histograms of \emph{lookahead} and \emph{number of branches} metrics for \texttt{ConceptNet}, \texttt{ogbl-wikikg2}, \texttt{ogbn-papers100M}, \texttt{ogb}, and \texttt{NaturalProofs}. There is no direct measure for number of branches for \texttt{NaturalProofs} since it is a dataset of proofs rather than graphs. The top row is lookaheads in log scale and the bottom row is number of branches in log scale. The red lines represent the $50^{\footnotesize th}$, $75^{\footnotesize th}$, $90^{\footnotesize th}$, $99^{\footnotesize th}$, and $99.9^{\footnotesize th}$ percentiles respectively.}
    \label{fig:lookahead_branch_histogram}
    \vspace{-1.3em}
\end{figure*}

Figure~\ref{natural_language_metrics} shows that the characteristic performance cliffs observed in the graph connectivity setting persist in the proof planning setting, and indeed the cliffs appear at lower complexities. Between lookaheads $L=10$ and $L=64$, \texttt{R1} and \texttt{o3} accuracies start collapsing for branch size $B>1$, though they still collapse at larger lookaheads than \texttt{V3} and \texttt{4o}. Although accuracy for $B=2$ and $B=4$ cases seems to stay above zero even as lookahead increases, this residual performance is not indicative of genuine reasoning ability. Rather, for a given number of branches $B$, the model can resort to a uniform random guess, achieving success with probability $1/B$. For $B=2$ we can see the performance hovering around $\frac{1}{2}$, and for $B=4$, we can see the performance hovering around $\frac{1}{4}$. Empirically, models that produce fewer ``thinking'' tokens tend to have non-zero accuracy for branches 2 and 4, and a somewhat extended plateau for branches of size 8 and 16. In fact, we can also observe this $1/B$ accuracy trend for \texttt{4o} in the symbolic case for $B=2$ (Figure~\ref{fig:symbolic_first_step} in Section~\ref{app:additional_analysis}). The full \texttt{o3} model also seems to converge to chance when run on lookaheads $L=300$ and $L=512$ and number of branches $B\in\{2,4,8,16\}$ (Figure \ref{natural_language_metrics}) despite using a large number of tokens (Figure~\ref{fig:symbolic_completion_tokens} in Section~\ref{app:additional_analysis}). It is difficult to perform deeper analysis since we do not have access to the reasoning tokens of \texttt{o3}. This behavior reflects stochastic guessing in LRMs rather than enhanced reasoning capabilities.

Among the models evaluated, \texttt{4o} attains the highest apparent accuracy solely by emitting very concise outputs---fewer than ten tokens on average---which we hypothesize leads the model to perform uniform random guessing more often and attain $1/B$ accuracy. By contrast, \texttt{V3}, despite also not being trained with reasoning incentives, generates verbose justifications that may discourage it from performing random guessing (Figure~\ref{natural_language_metrics} in Section~\ref{app:additional_analysis}). The reason for \texttt{V3}'s markedly higher token usage in the natural-language setting, relative to its behavior for symbolic graphs (Figure~\ref{fig:token_usage_graph}), remains unclear.

\paragraph{Verification of Real-World Proofs}
We evaluate the LRMs' ability to find errors and validate proofs using the \texttt{NaturalProofs} dataset. We structure this problem to evaluate the model's ability to (1) traverse the in-context proof DAG, (2) verify that each step is properly licensed by its premises, and (3) confirm that the conclusion is supported (i.e., the path is valid end-to-end), all in a natural reasoning setting. We first sample proofs via stratified sampling over proof length, ensuring balanced coverage across short, medium, and long proofs. For each sampled proof, we randomly sample a line in the proof to introduce a minimal but semantics-altering edit while preserving surface form (e.g., changing a quantifier, negation, inequality direction, number/variable/name). The modified proof is then given to the model with full proof context and is prompted to (1) judge whether the proof is correct and (2) if incorrect, identify the erroneous line. We note that, because this dataset is real-world natural language, it is quite noisy.
We calculate accuracy of error finding and report the accuracy along the axis of proof length (Figure~\ref{fig:proof_verfication}). We also carry out the trivial experiment of finding errors in unmodified proofs and see that the model performance can still suffer due to dataset noise or erroneous error identification (Figure~\ref{fig:proof_verification_trivial}). This suggests that our perturbation procedure is not the sole driver of the collapse. Looking at the metrics, we find patterns mirroring performance on synthetic graphs---namely, the performance abruptly drops at larger depths. Perturbation examples and proofs can be found in Table~\ref{tab:proof-perturbation-examples} of Section~\ref{app:proof_verification}.

\subsection{Inspection of Failure Modes}

We wish to gain insight into the failure modes of the models, by manually inspecting the ``thinking tokens'' of \texttt{R1} (such tokens are unavailable for \texttt{o3} and \texttt{o3-mini}). To that end, we sampled a number of symbolic graphs with $L \geq 300$ and $B=3$ for which \texttt{R1} returned no path and a number of symbolic graphs for which \texttt{R1} returned an incorrect path, we manually inspected the thinking tokens with those inputs, and categorized the mistakes according to a number of error types:
\begin{itemize}[noitemsep,topsep=0pt,leftmargin=1.6em]
    \item[I.] The model omits a necessary and previously-stated outgoing edge at some intermediate step (this occurred in 15/20 samples). Example of thinking trace: \textit{``Similarly, from 860: (860, ?) — I see (860, 227) ? (860, 227) is not listed''}.
    \item[II.] The model misses one of two outgoing edges of the start node, thus ignoring one of two branches entirely (5/20 samples). Example of thinking trace: \textit{``First, find all edges starting from 220. Looking at the list, I see (220, 723). That's the only one with 220 as the start.''}
    \item[III.] The model hallucinates an edge at some intermediate step (2/20 samples).
\end{itemize}
Under type I and type II errors, after the graph misses the required outgoing edge, thus pruning the gold branch, the model proceeds correctly on the wrong branch, and subsequent steps are locally consistent, leading to a “no path” response. For the examples for which the model returned an incorrect path, we find that Type III errors occurred most frequently (13/20 for Type I, 3/20 for Type II, 16/20 for Type III). In this case, the model often makes Type I and Type III errors simultaneously (11/20 cases) to derive a path which travels through the wrong branch, thus leading to an incorrect path result. These main failure modes from our analysis aligns with the notion of ``propagation error'' in \citet{dziri2023faithfatelimitstransformers}: after a single early local misstep, later computations are correct, conditioned on incorrect parent values. Their analysis found a similar conclusion of such “propagation errors” contributing to the decrease in the proportion of “fully correct” answers as depth increases.

\section{Complexity of Real-World Reasoning}

In this section, we aim to relate LRM behavior to the structural complexity of graphs and proofs that occur in real-world scenarios. This connection allows us to estimate where LRMs will succeed (the ``head'' of the distribution of lookahead and number of branches) and where they will predictably fail (the ``tail'') on real-world reasoning problems.

\paragraph{Real-World Datasets}
We analyze a diverse suite of large, real-world graphs. Most are drawn from the Open Graph Benchmark (OGB; \citealp{hu2021opengraphbenchmarkdatasets}), which includes code ASTs, academic citation networks, biomedical interaction graphs, and more. We also include \texttt{ConceptNet} \citep{speer2018conceptnet55openmultilingual}, a single integrated knowledge graph compiled from multiple sources, and the \texttt{NaturalProofs} corpus of formal proofs \citep{welleck2021naturalproofsmathematicaltheoremproving}.
We analyze the distributions of \texttt{ConceptNet}, \texttt{ogbl-wikikg2}, \texttt{ogbn-papers100M}, and \texttt{NaturalProofs} independently; and aggregate the other 24 OGB datasets into a single distribution (normalized uniformly across datasets) called \texttt{ogb}. 

\paragraph{Measures}
For each dataset, we compute (i) the lookahead from every node to every other node using an efficient algorithm (Algorithm~\ref{alg:bfs_lookahead} in Section~\ref{app:lookahead}). When this is computationally infeasible, we uniformly sample a large set of source nodes (e.g., 2M for \texttt{ConceptNet} and 30k for \texttt{ogbn-papers100M}) and compute lookaheads to all nodes. We also compute (ii) the number of branches (i.e., out-degree) of all nodes. For proofs such as those in \texttt{NaturalProofs}, the proof length (in terms of the number of steps) serves as a measure of complexity analogous to lookahead. This is due to the fact that for a sufficiently powerful formal system (e.g., including deduction rules for quantifiers, or arithmetic), there will often exist logical forms for which an infinite number of deductions are possible (e.g., given $x=y$, we can conclude $xz=yz$ for any $z$; or given $\forall x.f(x)$, we can conclude $f(a)$ for any term $a$). Thus, the number of ``branches'' at many nodes in a proof is infinite, and for any proof consisting of $n$ steps, there will almost always exist a disjoint proof with a different goal of unbounded length (i.e., sharing no proof steps with the original proof). In \texttt{NaturalProofs}, the proofs are given as a list of lines (in \texttt{Lean} or unstructured natural language) and we treat the number of lines as a proxy for proof length, which is analogous to lookahead.

\paragraph{Summary statistics}
As seen in Figure \ref{fig:lookahead_branch_histogram}, if each pair of nodes in a graph is taken as a graph connectivity example, the vast majority of examples seen in empirical graphs have quite low complexity. In our model evaluations for reasoning on graph connectivity (Figure \ref{fig:lookahead_metrics_4_branches}), we see the accuracy drop at around $L=100$ through $200$---sooner for high $B$. For \texttt{ogbl-wikikg2} and \texttt{ogb}, $L=100$ examples are in the $99^{\footnotesize th}$ percentile. Similarly, we see the LRM performance on next step prediction for proof planning collapses around $L=16$ through $32$ (Figure \ref{natural_language_metrics}) and we see that for \texttt{NaturalProofs}, $L=16$ and $L=32$ fall at the $50^{\footnotesize th}$ and $75^{\footnotesize th}$ percentile respectively. We also see that for all datasets, examples with number of branches $B=7$ fall in the $75^{\footnotesize th}$ percentile and there are datasets with branch counts in the thousands---well outside of the success regime of LRMs.

\section{Discussion}

We began by comparing LLMs and LRMs on the \texttt{NLGraph} benchmark, where LRMs outperformed LLMs---but only on low‑complexity graphs. We tested the models on examples with larger lookaheads and number of branches, and found that every model suffered an abrupt drop in accuracy. This performance cliff persisted even in examples with chain graphs (no search or backtracking), demonstrating that limitations in reasoning---not token limits or truncation---drive the collapse.

Extending this to natural language proof planning yielded the same result: LRM performance collapses as proof complexity increases, and in fact, the performance cliff appears at even smaller lookaheads than in the graph connectivity setting.

Finally, by measuring lookahead and branching distributions over real‑world graph and proof corpora, we showed that although most queries lie within the LRMs' success regime, a long tail of harder instances falls squarely beyond LRM capabilities. This mirrors prior work showing steep drops when test complexity exceeds training complexity \citep{dziri2023faithfatelimitstransformers,lee2025selfimprovingtransformersovercomeeasytohard}, and suggests that current LRMs remain tightly bound to their training distribution rather than exhibiting robust, human‑like generalization.

\section{Conclusion}
Current state-of-the-art LRMs excel on the relatively simple reasoning tasks that comprise much of existing benchmarks and real-world datasets, but falter once tasks become sufficiently complex. \texttt{DeepRD} enables exploration of the limits of LRM reasoning ability for both graph connectivity and proof planning, revealing abrupt failures that likewise appear in the long-tail of real-world reasoning tasks. Bridging this gap will require new methods to encourage more robust out-of-distribution generalization in reasoning.

\texttt{DeepRD} enables several interesting future research directions. First is a deeper analysis \texttt{R1}'s ``thinking'' tokens for traversal and reasoning: Are there any thinking patterns that are predictive of the model's performance? Another, enabled by unlimited data generation, is to test whether fine-tuning models on higher complexities in symbolic cases transfer performance to natural language reasoning.

\section*{Limitations}
One of our primary limitations is lack of open access to OpenAI's \texttt{o3} and \texttt{o3-mini}, or their training details. Running qualitative analysis on the thinking tokens of the OpenAI models would give a better idea of consistent failure modes across all LRMs. Also, given the high cost to run inference on state-of-the-art LLMs, and especially LRMs due to increase in token usage, another limitation is running the evaluation on more test cases. Currently we have a set of lookaheads and number of branch pairs we evaluate on, and that set can be made much larger with more evaluation budget. The budget also limited us to graph connectivity and next step prediction, but our methodology can be applied to other graph and reasoning problems. Finally, open access or even details for the training data of the LLMs and LRMs (for both DeepSeek and OpenAI) would allow us to analyze the complexity of the complete training distribution.

\section*{Acknowledgments}
We thank the anonymous reviewers for their considerate and helpful feedback. This work was supported in part by the Rosen Center for Advanced Computing (RCAC) resources, services, and staff expertise at Purdue University. We also thank the Department of Linguistics at the University of Washington for their guidance, support, and resources.

\bibliography{custom}

\appendix

\section{Appendix}

\raggedbottom

\subsection{Model Settings} \label{appendix_modelsettings}

\subsubsection{System Prompt}
The following system prompt was used for all models:
\begin{quote}
``When answering a yes or no question, always start the answer with |YES| or |NO| depending on your answer.''
\end{quote}
All models were run with a temperature setting of 0, except for the OpenAI reasoning models where the API does not allow customization of the temperature.

\subsubsection{Models Used}
The experiments utilized the following models:

\paragraph{From Together.ai:}
We used the API from \url{https://www.together.ai/} for the two models:
\begin{itemize}[itemsep=0pt,topsep=3pt]
    \item \texttt{deepseek-ai/DeepSeek-R1} (version 0528); 685B parameters
    \item \texttt{deepseek-ai/DeepSeek-V3} (version 0324); 671B parameters
\end{itemize}

\paragraph{From OpenAI:}
We used the API from \url{https://platform.openai.com/} for the four models below. We have no knowledge of any official model sizes (i.e., number of parameters).
\begin{itemize}[itemsep=0pt,topsep=3pt]
    \item \texttt{gpt-4o-mini-2024-07-18} (for extracting paths from the final answer)
    \item \texttt{o3-mini-2025-01-31}
    \item \texttt{o3-2025-04-16}
    \item \texttt{gpt-4o-2024-08-06}
\end{itemize}

\subsection{Computing Lookahead}\label{app:lookahead}

Given a start vertex $s$ and a goal vertex $g$ in a directed graph, we define \emph{lookahead} as the number of breadth-first search (BFS) iterations needed to determine the next vertex along a path from $s$ to $g$. Alg.~\ref{alg:bfs} provides pseudocode for a simple BFS implementation.

We can minimally modify the BFS algorithm to compute the lookahead, as described above, which is shown in Alg.~\ref{alg:lookahead}. Intuitively, we essentially perform BFS until we either: (1) find the goal $g$, or (2) determine that any path from $s$ to $g$ must pass through a child of $s$. Alg.~\ref{alg:lookahead} keeps track of, for each node in the frontier, which child of $s$ can reach it; as soon as all surviving paths to $g$ must pass through the same child of $s$, we stop and return that layer index $L$ (the required lookahead). We use this to compute lookahead, our main measure of graph complexity, as explained in Section~\ref{sec:complexity}.

In our analysis, we need to be able to compute the lookahead between many pairs of nodes in a given graph. While applying Alg.~\ref{alg:lookahead} for each pair is a valid approach, it can be costly in terms of running time. Therefore, we can modify the algorithm to compute the lookahead from a source $s$ to all possible targets. We observe that Alg.~\ref{alg:lookahead} only depends on the target $g$ at the first step within each iteration, and so we can modify this part of the algorithm to compute the lookahead to all possible targets, which is shown in Alg.~\ref{alg:bfs_lookahead}.

In our analysis, we similarly need to compute the all-pairs shortest-path lengths for a given graph. We implement a simple approach to do so in Alg.~\ref{alg:pairwise_distance}. This algorithm essentially runs BFS iteratively for all pairs of vertices.

\begin{algorithm}
\footnotesize
\let\oldnl\nl
\newcommand{\nonl}{\renewcommand{\nl}{\let\nl\oldnl}}
\SetNlSty{}{\color{RedOrange}\sffamily}{}
\SetAlgoBlockMarkers{}{}
\SetKwProg{Fn}{function}{}{}
\SetKwIF{If}{ElseIf}{Else}{if}{ }{else if}{else }{}
\SetKw{Continue}{continue}
\SetKwFunction{FDistance}{\small distance}
\SetKwFor{For}{for}{do}{end}
\SetKwFor{While}{while}{do}{end}
\SetKwProg{uForEach}{for each}{ do}{}
\SetKwProg{Fn}{function}{}{}
\AlgoDisplayBlockMarkers\SetAlgoVlined
\SetAlCapNameFnt{\small}
\SetAlCapFnt{\small}
\SetNoFillComment
\DontPrintSemicolon
\SetInd{0.0em}{0.8em}
    \Fn{\FDistance{source node $s$, goal node $g$}}{
        \lIf{$s = g$}{
            \Return{0}
        }
        initialize $d = 1$ \;
        initialize $Q = \{c : c\text{ is a child node of }s\}$ \;
        \While(\tcc*[f]{BFS main loop}){$Q$ is not empty}{
            \lIf{$g \in Q$}{
                \Return{$d$}
            }
            set $Q = \{c : c\text{ an \emph{unvisited} child of any }v\in Q\}$ \;
            set $d = d + 1$ \;
        }
        \Return{$\infty$}
	}
	\caption{Breadth-first search to compute the distance from node $s$ to node $g$.}
	\label{alg:bfs}
\end{algorithm}

\begin{algorithm}
\footnotesize
\let\oldnl\nl
\newcommand{\nonl}{\renewcommand{\nl}{\let\nl\oldnl}}
\SetNlSty{}{\color{RedOrange}\sffamily}{}
\SetAlgoBlockMarkers{}{}
\SetKwProg{Fn}{function}{}{}
\SetKwIF{If}{ElseIf}{Else}{if}{ }{else if}{else }{}
\SetKw{Continue}{continue}
\SetKwFunction{FLookahead}{\small lookahead}
\SetKwFor{For}{for}{do}{end}
\SetKwFor{While}{while}{do}{end}
\SetKwProg{uForEach}{for each}{ do}{}
\SetKwProg{Fn}{function}{}{}
\AlgoDisplayBlockMarkers\SetAlgoVlined
\SetAlCapNameFnt{\small}
\SetAlCapFnt{\small}
\SetNoFillComment
\DontPrintSemicolon
\SetInd{0.0em}{0.8em}
    \Fn{\FLookahead{source node $s$, goal node $g$}}{
        \lIf{$s = g$}{
            \Return{0}
        }
        initialize $l = 1$ \;
        initialize $R_v = \varnothing$ for all nodes $v$ \;
        initialize $Q = \{c : c\text{ is a child node of }s\}$ \;
        set $R_c = \{c\}$ for every child node $c$ of $s$ \;
        \While(\tcc*[f]{BFS main loop}){$Q$ is not empty}{
            \lIf{$g \in Q$}{
                \Return{$l$}
            }
            \ElseIf{$c \in R_v$ for all $v \in Q$ for some child $c$ of $s$}{
                \Return{$l$}
            }
            \For{$v \in Q$ and every child node $c$ of $v$} {
                set $R_c = R_c \cup R_v$ \;
            }
            set $Q = \{c : c\text{ an \emph{unvisited} child of any }v\in Q\}$ \;
            set $l = l + 1$ \;
        }
        \Return{$\infty$}
	}
	\caption{Algorithm to compute the lookahead from node $s$ to node $g$.}
	\label{alg:lookahead}
\end{algorithm}

\begin{algorithm}
\footnotesize
\let\oldnl\nl
\newcommand{\nonl}{\renewcommand{\nl}{\let\nl\oldnl}}
\SetNlSty{}{\color{RedOrange}\sffamily}{}
\SetAlgoBlockMarkers{}{}
\SetKwProg{Fn}{function}{}{}
\SetKwIF{If}{ElseIf}{Else}{if}{ }{else if}{else }{}
\SetKw{Continue}{continue}
\SetKwFunction{FLookaheads}{\small lookaheads\_from\_src}
\SetKwFor{For}{for}{do}{end}
\SetKwFor{While}{while}{do}{end}
\SetKwProg{uForEach}{for each}{ do}{}
\SetKwProg{Fn}{function}{}{}
\AlgoDisplayBlockMarkers\SetAlgoVlined
\SetAlCapNameFnt{\small}
\SetAlCapFnt{\small}
\SetNoFillComment
\DontPrintSemicolon
\SetInd{0.0em}{0.8em}
    \Fn{\FLookaheads{source node $s$}}{

        initialize $l = 1$ \;
        initialize $L_v = \infty$ for all nodes $v$\;
        initialize $R_v = \varnothing$ for all nodes $v$\;
        initialize $Q = \{c \mid c\text{ is a child of }s\}$\;
        set $R_c = \{c\}$ for every child node $c$ of $s$\;

        \While(\tcc*[f]{BFS main loop}){$Q$ is not empty} {
            \If{$c \in R_v$ for all $v \in Q$ for some child $c$ of $s$} {
                \textbf{break} \;
            }
            \For{$v \in Q$ and every child node $c$ of $v$} {
                set $R_c = R_c \cup R_v$ \;
            }
            set $L_v = l$ for all $v \in Q$ \;
            set $Q = \{c : c\text{ an \emph{unvisited} child of any }v\in Q\}$ \;
            set $l = l + 1$ \;
        }

        \ForEach{$v$ is any descendant of a node in $Q$}{
            set $L_v = l$ \;
        }

        \Return{$L$} \;
    }
\caption{Computes the lookaheads from a given source node $s$ to all other nodes in the graph. In our experiments, we add an additional loop over source nodes $s$, in order to compute the lookaheads between \emph{all pairs} of nodes.}
\label{alg:bfs_lookahead}
\end{algorithm}

\begin{algorithm}
\footnotesize
\let\oldnl\nl
\newcommand{\nonl}{\renewcommand{\nl}{\let\nl\oldnl}}
\SetNlSty{}{\color{RedOrange}\sffamily}{}
\SetAlgoBlockMarkers{}{}
\SetKwProg{Fn}{function}{}{}
\SetKwIF{If}{ElseIf}{Else}{if}{ }{else if}{else }{}
\SetKw{Continue}{continue}
\SetKwFunction{PairwiseDistance}{\small pairwise\_distance}
\SetKwFor{For}{for}{do}{end}
\SetKwFor{While}{while}{do}{end}
\SetKwProg{uForEach}{for each}{ do}{}
\SetKwProg{Fn}{function}{}{}
\AlgoDisplayBlockMarkers\SetAlgoVlined
\SetAlCapNameFnt{\small}
\SetAlCapFnt{\small}
\SetNoFillComment
\DontPrintSemicolon
\SetInd{0.0em}{0.8em}
\Fn{\PairwiseDistance{graph $G$}}{
    initialize $\mathit{Dist} =$ empty counter\;
    \tcc*[f]{distribution of pairwise shortest-path lengths}\\[2pt]
    
    \ForEach{node $s$ in $G$}{
        initialize $\mathit{visited} = \{s\}$\;
        initialize $Q = \{c \mid c \text{ is a child of } s\}$\;
        initialize $d = 1$\;
        
        \While(\tcc*[f]{BFS main loop}){$Q$ is not empty}{
            \ForEach{$v \in Q$}{
                add one occurrence of $d$ to $\mathit{Dist}$\;
                add $v$ to $\mathit{visited}$\;
            }
            set $Q = \{c \mid c \text{ is a child of any } v \in Q \text{ and } c \notin \mathit{visited}\}$\;
            set $d = d + 1$\;
        }
    }
    \Return{$\mathit{Dist}$}\;
}
\caption{Computes the distribution of shortest-path distances between all pairs of nodes in $G$.}
\label{alg:pairwise_distance}
\end{algorithm}

\subsection{Translating from Symbolic Query to Proof Planning}
\label{appendix:translation}

\begin{algorithm}
\footnotesize
\let\oldnl\nl
\newcommand{\nonl}{\renewcommand{\nl}{\let\nl\oldnl}}
\SetNlSty{}{\color{RedOrange}\sffamily}{}
\SetAlgoBlockMarkers{}{}
\SetKwProg{Fn}{function}{}{}
\SetKwIF{If}{ElseIf}{Else}{if}{ }{else if}{else }{}
\SetKw{Continue}{continue}
\SetKwFunction{FTranslate}{\small graph\_to\_logic}
\SetKwFor{For}{for}{do}{end}
\SetKwFor{While}{while}{do}{end}
\SetKwProg{uForEach}{for each}{ do}{}
\SetKwProg{Fn}{function}{}{}
\AlgoDisplayBlockMarkers\SetAlgoVlined
\SetAlCapNameFnt{\small}
\SetAlCapFnt{\small}
\SetNoFillComment
\DontPrintSemicolon
\SetInd{0.0em}{0.8em}

\Fn{\FTranslate{edge list $E = [(A,B)]$, query $(X,Y)$, where $A,B,X,Y$ are integers representing the nodes}}{

    initialize $V =$ unique nodes appearing in $E$\;
    initialize $n = |V|$\;
    
    initialize $name = $ a random name\;
    initialize $Adjs  = $ list of $n$ distinct random attributes\;
    \tcc*[f]{attributes are random two-syllable words}
    
    sort $V$ into $[v_0,\dots,v_{n-1}]$\;

    \nonl\begin{minipage}[t]{\linewidth}
    \begin{flushleft}
    {\small
    initialize $Choices =$ array of eight strings:\par
    \quad 1. ``If $name$ is {$adj\_A$}, then $name$ is {$adj\_B$}.''\par
    \quad 2. ``$name$ is {$adj\_A$} implies $name$ is {$adj\_B$}.''\par
    \quad 3. ``$name$ is {$adj\_B$} is true if $name$ is {$adj\_A$}.''\par
    \quad 4. ``{$adj\_B$} is true if {$adj\_A$} is true.''\par
    \quad 5. ``If {$adj\_A$} then {$adj\_B$} is true.''\par
    \quad 6. ``If {$adj\_A$} is true then {$adj\_B$}.''\par
    \quad 7. ``Given $name$ is {$adj\_A$} then $name$ is {$adj\_B$}.''\par
    \quad 8. ``If a person is {$adj\_A$} then they are {$adj\_B$}.''}
    \end{flushleft}
    \end{minipage}
    
    initialize $Preds = \varnothing$\;
    
    \ForEach{$(A,B) \in E$} {
        set $adj_A = Adjs[A]$\;
        set $adj_B = Adjs[B]$\;
        initialize $predicate = $ random string from $Choices$.\;
        append $predicate$ to $Preds$\;
    }
    
    initialize $adj_X = Map[X]$,\;
    initialize $adj_Y = Map[Y]$\;
    $Q \gets$ ``Given that \textit{name} is \textit{adj\_X}, and we want to prove
               \textit{name} is \textit{adj\_Y}. The next step in the proof is:
               \textit{name} is \underline{\phantom{xxx}}.''\;
               
    \Return{$Preds,\;Q$}\;
}
\caption{Translate an edge list and a query \((X,Y)\) into natural-language predicates and a fill-in-the-blank logic question.}
\label{alg:graph_to_puzzle}
\end{algorithm}

We wish to translate the graph search query from symbolic to natural language proof planning. We do this by first choosing a random name (e.g., \textit{Alyssa}), which we use as the subject for the natural language query. We then assign each node a unique attribute, which is a fake two-syllable word. Each syllable consists of one of the following forms: Consonant (C) + Vowel (V), VC, CVC, CVV, CCV, VCV, or VCC. Examples of these attributes are "\textit{movo}", or "\textit{retil}", etc. Then each directed edge is turned into a predicate involving two attributes, such as "\textit{If Alyssa is movo, then Alyssa is retil} or \textit{Alyssa is movo implies Alyssa is retil}", etc. We use eight different predicate forms to introduce linguistic variety. The query itself, consisting of a start and goal node, is also translated into natural language. The pseudocode which describes this process is provided in Alg.~\ref{alg:graph_to_puzzle}.

\subsection{Graph Sampling}
\label{appendix:sampling}

\begin{algorithm}
\footnotesize
\let\oldnl\nl
\newcommand{\nonl}{\renewcommand{\nl}{\let\nl\oldnl}}
\SetNlSty{}{\color{RedOrange}\sffamily}{}
\SetAlgoBlockMarkers{}{}
\SetKwProg{Fn}{function}{}{}
\SetKwIF{If}{ElseIf}{Else}{if}{ }{else if}{else}{}
\SetKw{Continue}{continue}
\SetKw{Break}{break}
\SetKwFunction{FGenGraph}{\small generate\_lookahead\_graphs}
\SetKwFunction{FLookahead}{\small compute\_lookahead}
\SetKwFunction{FPrefAttach}{\small preferential\_attachment}
\SetKwFor{For}{for}{do}{end}
\SetKwFor{While}{while}{do}{end}
\AlgoDisplayBlockMarkers\SetAlgoVlined
\SetAlCapNameFnt{\small}
\SetAlCapFnt{\small}
\SetNoFillComment
\DontPrintSemicolon
\SetInd{0.0em}{0.8em}

\Fn{\FGenGraph{dataset size $N$, lookahead $l$, branches $b$, max edges $m$}}{
    initialize $\mathcal{D} = \varnothing$\;
    
    \While{$|\mathcal{D}| < N$}{
        set $n = \max(2, 1 + b \cdot l + \text{random}(0, m/3))$\;
        initialize vertices $V = \{v_0, \ldots, v_{n-1}\}$, edges $E = \varnothing$\;
        
        \tcc{Build main path}
        \For{$i = 0$ to $l - 1$}{
            $E \gets E \cup \{(v_i, v_{i+1})\}$\;
        }
        set start $s = v_0$, goal $g = v_l$\;
        
        \tcc{Add alternative branches}
        set $k = l + 1$\;
        \For{$j = 1$ to $b - 1$}{
            $E \gets E \cup \{(v_0, v_k)\}$\;
            set branch\_length $= l + \text{random}(0, \min(2, n - k))$\;
            \For{$i = 1$ to branch\_length $- 1$}{
                $E \gets E \cup \{(v_{k+i-1}, v_{k+i})\}$\;
            }
            $k \gets k +$ branch\_length\;
        }
        
        \tcc{Preferential attachment phase}
        initialize valid\_paths $\mathcal{V} =$ descendants of $v_1$\;
        \For{remaining vertices $v_{new}$}{
            \If{$|E| \geq m$}{\Break}
            
            sample $d_{in}, d_{out} \sim (d+1)^{-1}$\;
            
            select children $\mathcal{C} \subseteq V$ with \FPrefAttach{in-degree}\;
            connect $v_{new}$ to each child in $\mathcal{C}$\;
            
            compute valid parents avoiding cycles\;
            select parents $\mathcal{P} \subseteq V$ with \FPrefAttach{out-degree}\;
            connect each parent in $\mathcal{P}$ to $v_{new}$\;
            
            \If{\FLookahead{$s, v_1, g$} $\neq l$}{
                rollback edges for $v_{new}$\;
            }
        }
        
        \tcc{Finalization}
        remove excess edges while preserving paths from $s$ to $g$\;
        shuffle node IDs\;
        
        find shortest path $P$ from $s$ to $g$\;
        \If{$P$ does not exist or $|P| \leq 2$}{\Continue}
        
        \tcc{Validation}
        \If{out-degree$(s) \neq b$}{\Continue}
        \If{multiple simple paths exist from $s$ to $g$}{\Continue}  
        \If{graph isomorphic to any in $\mathcal{D}$}{\Continue}
        \If{exists path shorter than $l$ from $s$ to $g$}{\Continue}
        
        \tcc{Create single training example from first step}
        set $v_{curr} = P[0] = s$, $v_{next} = P[1]$\;
        \If{\FLookahead{$v_{curr}, v_{next}, g$} $= l$}{
            \If{only one child of $v_{curr}$ has path to $g$}{
                add $(E, s, g, v_{next})$ to $\mathcal{D}$\;
            }
        }
    }
    
    \Return{$\mathcal{D}$}\;
}
\caption{Generate graph dataset with controlled lookahead property for pathfinding tasks.}
\label{alg:generate_lookahead_graph}
\end{algorithm}

We use an algorithm (Alg.~\ref{alg:generate_lookahead_graph}) that generates directed acyclic graphs (DAGs) of a specific lookahead. We begin with constructing a base graph with $n$ nodes, where $n$ is chosen to have at least $b \text{ (branches) } \cdot l \text{ (lookahead) }$ nodes and additional random nodes. A main path of length $l$ is created from node $v_0$ to $v_l$, establishing the golden route from start to goal. Then, $b-1$ alternative branches are added, each starting from the start node $v_0$ and having length approximately equal to $l$ with slight random variation. This ensures that from the start node, there are exactly $b$ possible initial choices, but only by looking ahead $l$ steps can a model identify which choice leads to the goal.

The preferential attachment phase adds complexity to the graph while preserving the lookahead property. New nodes are connected with a power-law degree distribution so that the probability of having degree $d$ is proportional to $(d+1)^{-1}$, determining how many edges to add. For selecting which existing nodes to connect to, the algorithm uses preferential attachment: when adding outgoing edges from a new node, it selects target nodes with probability $P(v) \propto \alpha + \text{in-degree}(v)$, and when adding incoming edges, it selects source nodes with probability $P(v) \propto \alpha + \text{out-degree}(v)$. This creates a heavy-tailed degree distribution where most nodes are sparsely connected and a few nodes become densely connected hubs. The algorithm tracks which nodes lie on valid paths from start to goal and carefully avoids creating cycles. If adding edges to a new node would reduce the lookahead below $l$, those edges are rolled back.

After construction, the algorithm performs several validation steps. Node IDs are randomly shuffled to prevent models from learning spurious correlations between topology and numbering. The graph is checked to ensure: (1) exactly one simple path exists from start to goal when considering all edges, (2) the start node has exactly $b$ outgoing edges, (3) no path shorter than $l$ steps exists from start to goal, and (4) the graph structure is not isomorphic to previously generated graphs.

\subsection{Example Model Responses} \label{appendix_modelresponses}

Tables \ref{tab:v3-transcript} and \ref{tab:r1-transcript} show model outputs (reasoning + response) for sample proof planning \texttt{V3} and \texttt{R1} runs.

\begin{table}
\centering
\footnotesize 
\renewcommand\arraystretch{1.1}
\begin{tabular}{|>{\ttfamily}p{0.97\linewidth}|}
\hline
\texttt{V3} model response\\
\hrulefill\\
query:\\
Suppose we have the following list of facts:\\
If someone is eyzvih, they are tucibm.\\
If Mariah Gonzalez is eyzvih, then Mariah Gonzalez is ttaask.\\
If a person is veubte, they are uzvymi.\\
If Mariah Gonzalez is emkatv, then Mariah Gonzalez is rowzo.\\
If a person is uztio, they are otis.\\
Everyone that is rowzo is hatho.\\
\ldots\\
Given that Mariah Gonzalez is moilu, and we want to prove Mariah Gonzalez is otis. The next step in the proof is: Mariah Gonzalez is \_\_\_\_.\\
\hrulefill\\
Let's break down the steps to find out what Mariah Gonzalez is next, given that she is moilu and we want to prove she is otis.\\
\\
1. Given: Mariah Gonzalez is moilu.\\
2. From "If Mariah Gonzalez is moilu, then Mariah Gonzalez is zdedoa." $\rightarrow$ Mariah Gonzalez is zdedoa.\\
3. From "If someone is zdedoa, they are vupon." $\rightarrow$ Mariah Gonzalez is vupon.\\
4. From "If Mariah Gonzalez is vupon, then Mariah Gonzalez is xbaann." $\rightarrow$ Mariah Gonzalez is xbaann.\\
5. From "If someone is xbaann, they are gohsi." $\rightarrow$ Mariah Gonzalez is gohsi.\\
6. From "Everyone that is gohsi is uxfeyl." $\rightarrow$ Mariah Gonzalez is uxfeyl.\\
\textellipsis\\
29. From step 28, Mariah Gonzalez is uztio $\rightarrow$ she is otis (which is our goal).\\
\\
The next step after Mariah Gonzalez is moilu is that she is zdedoa (from step 2).\\
\\
Final Answer: zdedoa\\
\hline
\end{tabular}
\caption{\texttt{V3} model query and reasoning transcript for a given natural language prompt.}
\label{tab:v3-transcript}
\end{table}

\begin{table}
\centering
\footnotesize 
\renewcommand\arraystretch{1.1}
\begin{tabular}{|>{\ttfamily}p{0.97\linewidth}|}
\hline
\texttt{R1} model response\\
\hrulefill\\
query:\\
Suppose we have the following list of facts:\\
If someone is eyzvih, they are tucibm.\\
If Mariah Gonzalez is eyzvih, then Mariah Gonzalez is ttaask.\\
If a person is veubte, they are uzvymi.\\
If Mariah Gonzalez is emkatv, then Mariah Gonzalez is rowzo.\\
If a person is uztio, they are otis.\\
Everyone that is rowzo is hatho.\\
\ldots\\
Given that Mariah Gonzalez is moilu, and we want to prove Mariah Gonzalez is otis. The next step in the proof is: Mariah Gonzalez is \_\_\_\_.\\
\hrulefill\\
<think>\\
We are given that Mariah Gonzalez is moilu, and we want to prove that Mariah Gonzalez is otis.\\
 Let's list the facts that involve "moilu" and see what we can derive:\\
 1. "If Mariah Gonzalez is moilu, then Mariah Gonzalez is zdedoa."\\
    Since Mariah Gonzalez is moilu, then she is zdedoa.\\
 2. "If someone is zdedoa, they are vupon."\\
    So, from zdedoa, we get that Mariah Gonzalez is vupon.\\
 3. "If Mariah Gonzalez is vupon, then Mariah Gonzalez is xbaann."\\
    Therefore, Mariah Gonzalez is xbaann.\\
\textellipsis\\
Let's trace that:\\
    Given: moilu -> xori (from "Each person that is moilu is xori.")\\
    Then: xori -> emkatv (from "If Mariah Gonzalez is xori then they are emkatv.")\\
    Then: emkatv -> rowzo (from "If Mariah Gonzalez is emkatv, then Mariah Gonzalez is rowzo.")\\
    Then: rowzo -> hatho (from "Everyone that is rowzo is hatho.")\\
\textellipsis\\
</think>\\
Given that Mariah Gonzalez is moilu, the goal is to prove she is otis. The path to otis involves the following chain:\\
- From "Each person that is moilu is xori," since Mariah is moilu, she must be xori.\\
- From "If Mariah Gonzalez is xori then they are emkatv," she becomes emkatv.\\
- From "If Mariah Gonzalez is emkatv, then Mariah Gonzalez is rowzo," she becomes rowzo.\\
\textellipsis\\
Final Answer: xori\\
\hline
\end{tabular}
\caption{\texttt{R1} model query and reasoning transcript for a given natural language prompt.}
\label{tab:r1-transcript}
\end{table}

\subsection{Error Analysis}\label{app:error_analysis}

In this section, we provide additional results from analyzing the various errors we observe from the model. We categorize our analysis of these errors into symbolic and logic graphs with $B>1$, where we plot error rate vs lookahead, as well as symbolic and logic graphs with $B=1$, where we plot error rate vs graph depth, which is analogous to lookahead for single-branch (chain) graphs.

\begin{figure}
     \centering
     \includegraphics[width=\linewidth]{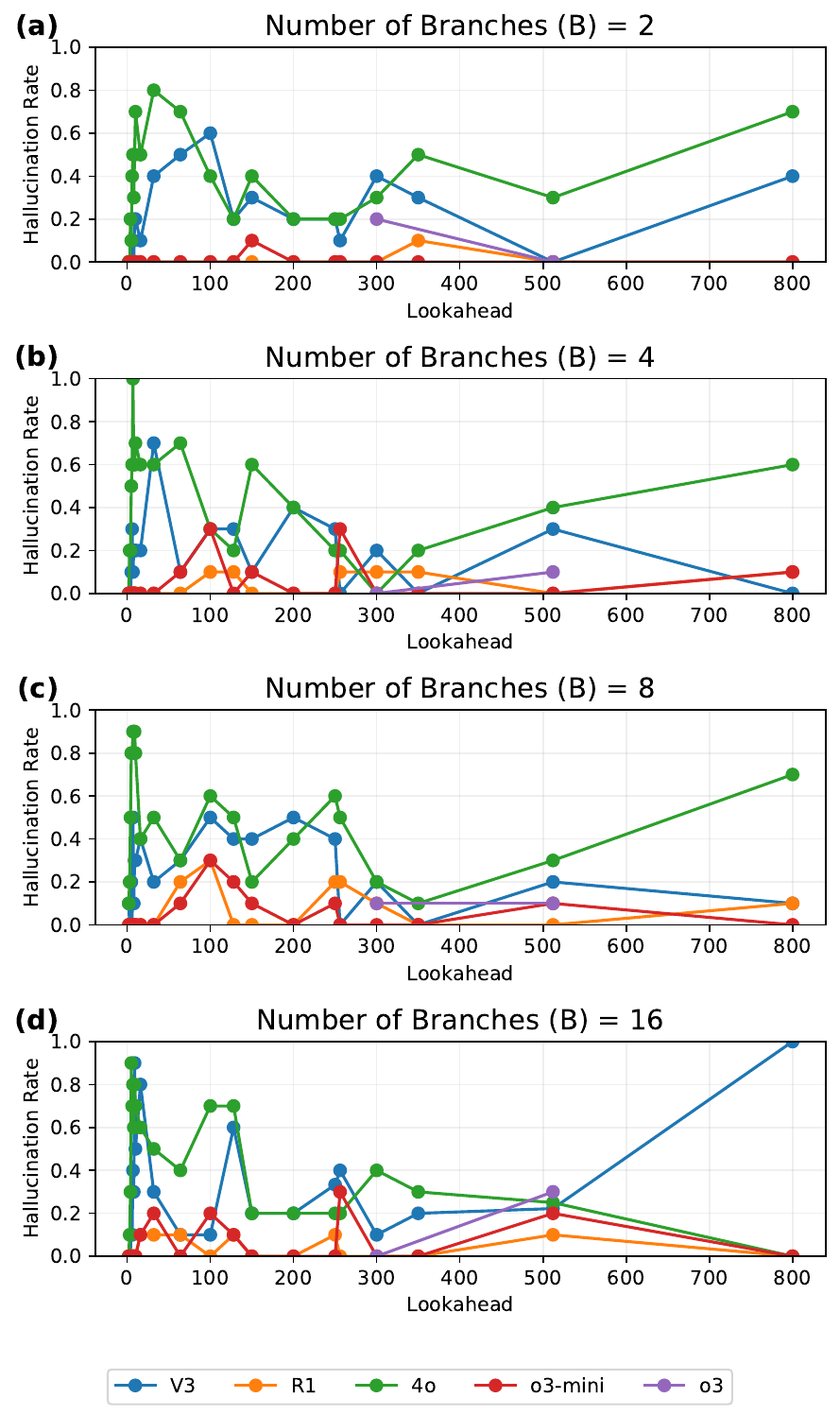}
     \caption{Edge hallucination rates for symbolic graph with $B > 1$.}
     \label{fig:symbolic_edge_hallucinations}
 \end{figure}

\begin{figure}
     \centering
     \includegraphics[width=\linewidth]{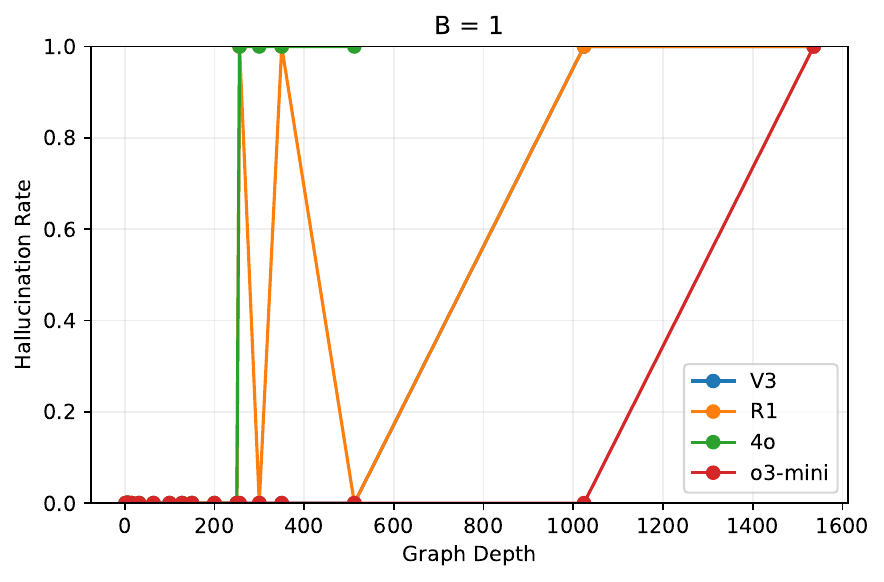}
     \caption{Edge hallucination rates for symbolic graph with $B = 1$.}
     \label{fig:symbolic_edge_hallucinations_branch1}
 \end{figure}

\paragraph{Edge Hallucination Rate}
An edge hallucination is defined as the appearance of an edge in the model's output that did not exist in the input symbolic graph. The edge hallucination rate tracks the proportion of runs per lookahead and branch size in which the model produced at least one edge hallucination. For $B>1$ (Fig.~\ref{fig:symbolic_edge_hallucinations}), rates are high at small lookaheads, drop to near zero toward medium lookaheads, then rise again at very high lookaheads. For $B=1$ (Fig.~\ref{fig:symbolic_edge_hallucinations_branch1}), rates stay near zero through moderate depths then rise sharply at large depths. The depth at which the rate begins to rise depends on the model.
 
\begin{figure}
     \centering
     \includegraphics[width=\linewidth]{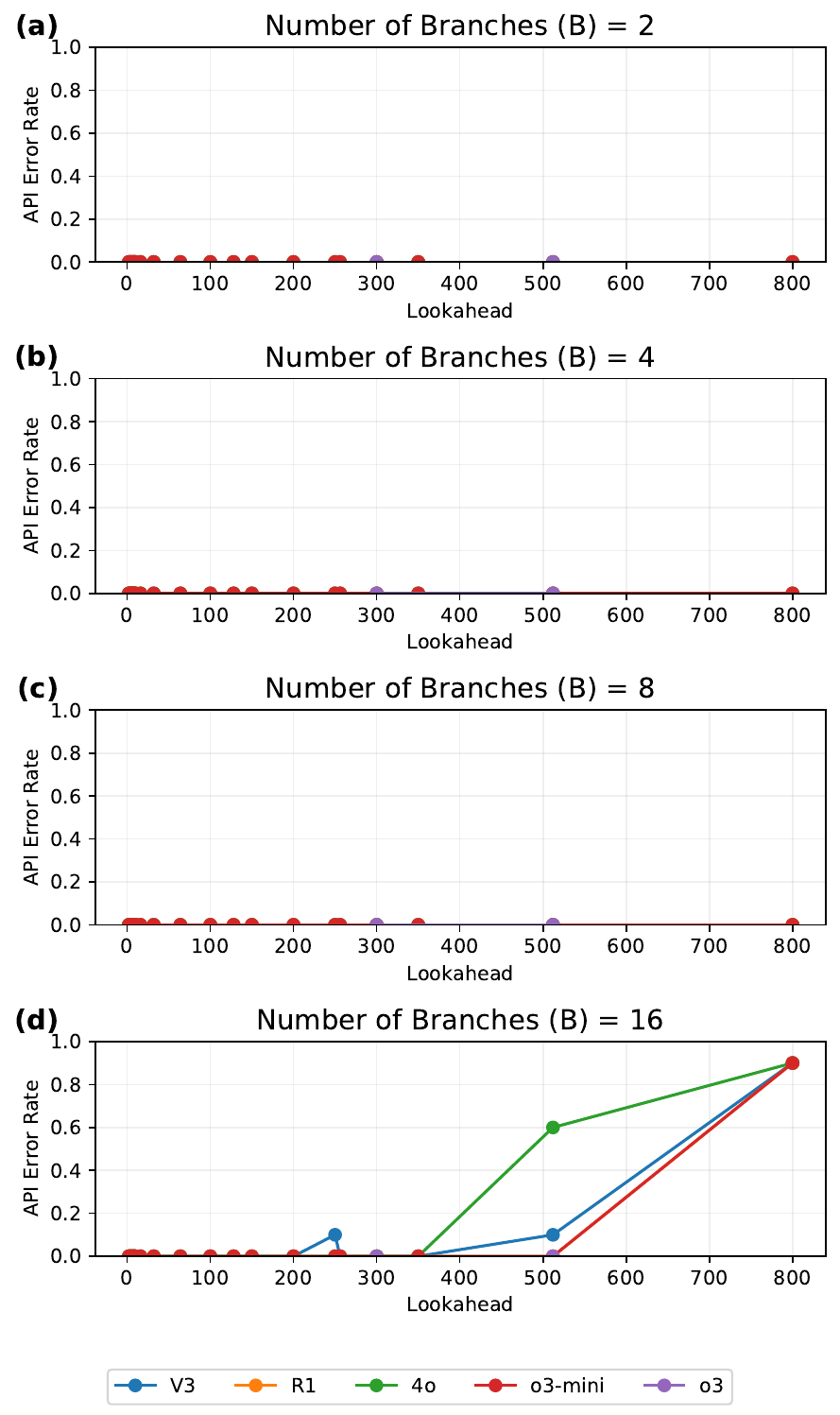}
     \caption{API error rate for symbolic graph with $B > 1$.}
     \label{fig:symbolic_api_errors}
 \end{figure}

\begin{figure}
     \centering
     \includegraphics[width=\linewidth]{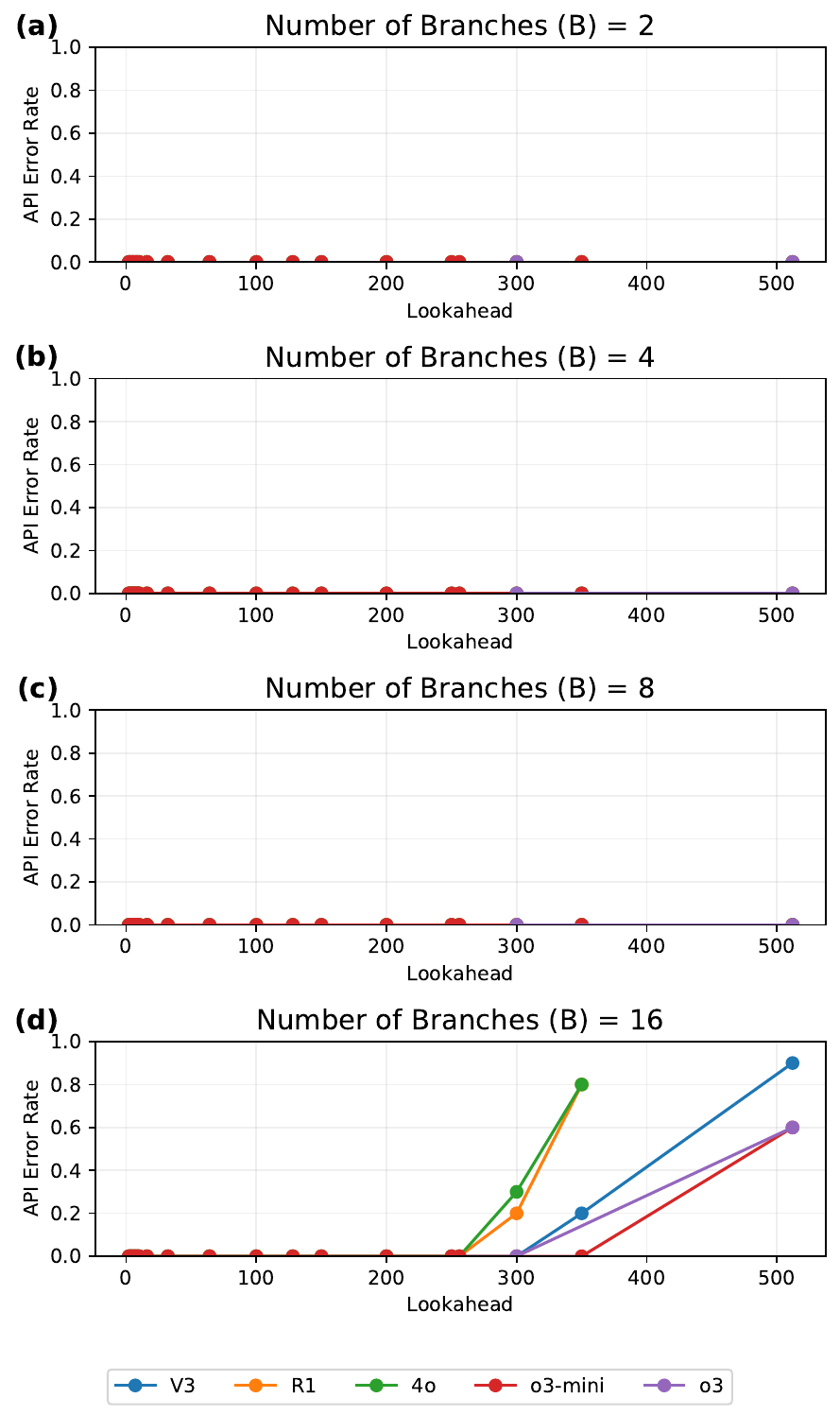}
     \caption{API error rate for logical graph with $B > 1$.}
     \label{fig:logic_api_errors}
 \end{figure}

\paragraph{API Error Rate}
A run is aborted by the API when the number of tokens needed to complete the prompt exceeds the amount supported by the API. Fig.~\ref{fig:symbolic_api_errors} and Fig.~\ref{fig:logic_api_errors} shows the proportion of these errors for the symbolic and logic graphs, respectively. In both cases, API errors only started appearing for $B=16$.

\begin{figure}
     \centering
     \includegraphics[width=\linewidth]{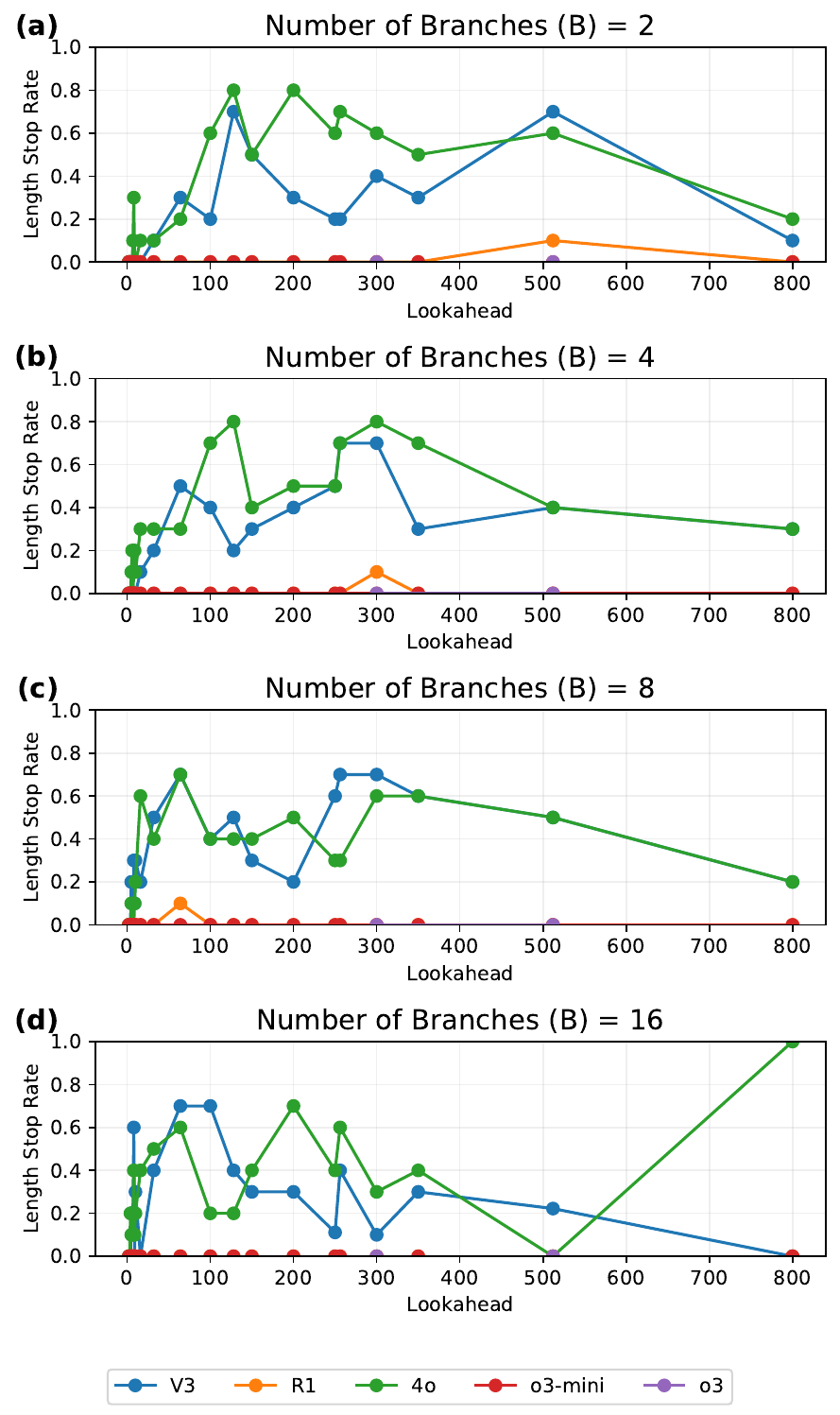}
     \caption{Length stop rate for symbolic graphs with $B > 1$.}
     \label{fig:symbolic_length_stops}
 \end{figure}

\begin{figure}
     \centering
     \includegraphics[width=\linewidth]{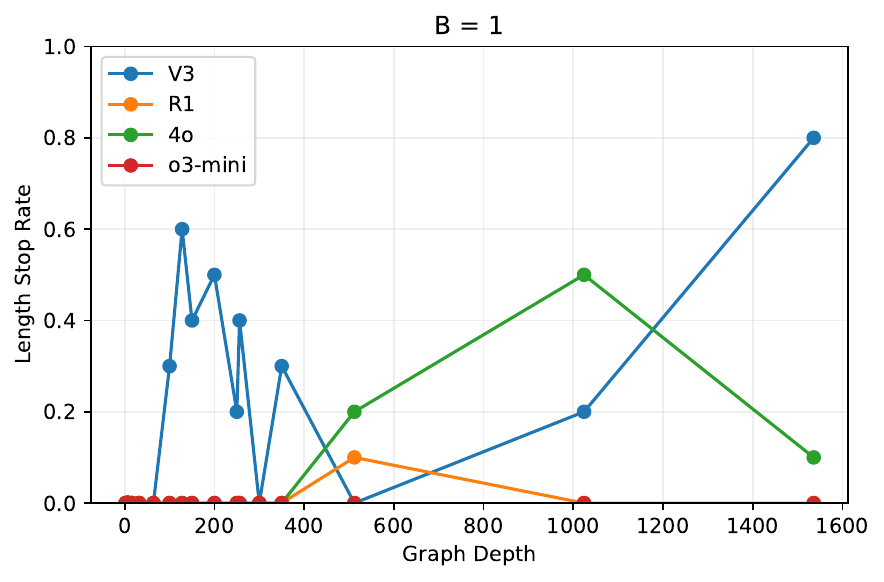}
     \caption{Length stop rate for symbolic graphs with $B = 1$.}
     \label{fig:symbolic_length_stops_branch1}
 \end{figure}

\begin{figure}
     \centering
     \includegraphics[width=\linewidth]{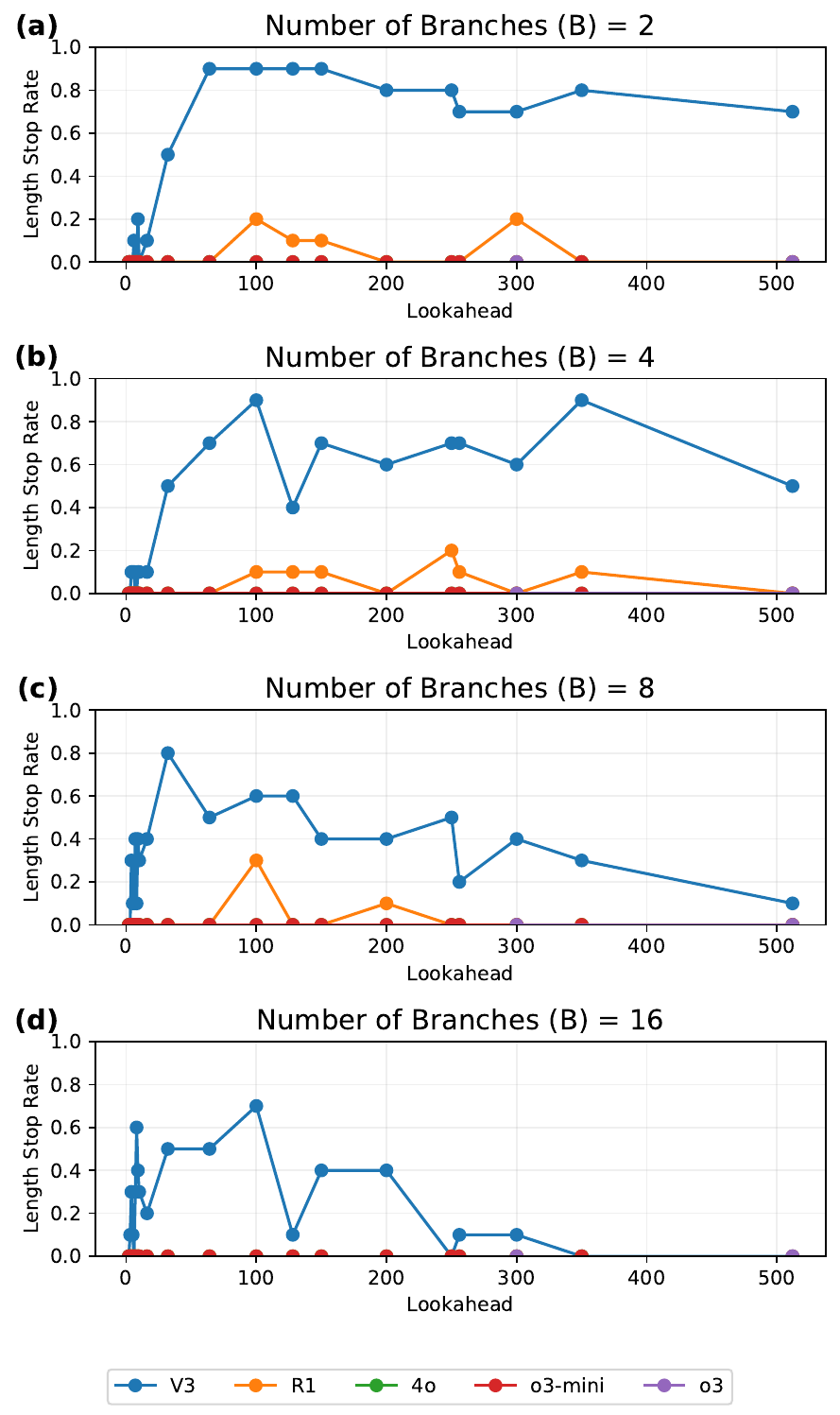}
     \caption{Length stop rate for logical graphs with $B > 1$.}
     \label{fig:logic_length_stops}
 \end{figure}

\begin{figure}
     \centering
     \includegraphics[width=\linewidth]{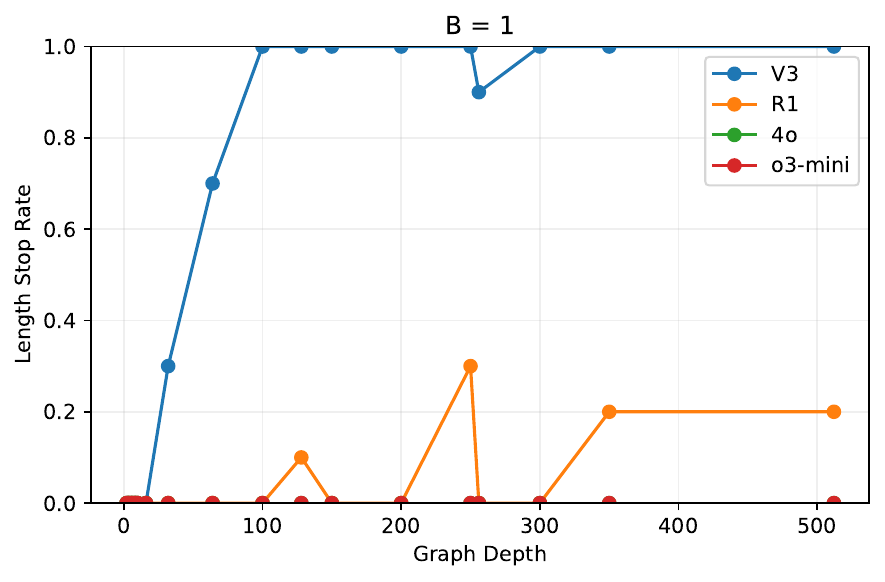}
     \caption{Length stop rate for logical graphs with $B = 1$.}
     \label{fig:logic_length_stops_branch1}
 \end{figure}

\paragraph{Length Stop Error}
The models will abruptly stop its processing of a prompt if the amount of tokens used during processing exceeds its limit. This rate tracks the proportion of runs per lookahead and branch size which was stopped due to reaching this length limit. Our observed rates largely correlate with the completion token usage, with a rise from small lookahead into a mid-lookahead peak and then taper at larger lookahead, and varying drastically by model. This behavior is consistently observed for symbolic graphs with $B>1$ (Fig.~\ref{fig:symbolic_length_stops}), symbolic graphs with $B=1$ (Fig.~\ref{fig:symbolic_length_stops_branch1}), logic graphs with $B>1$ (Fig.~\ref{fig:logic_length_stops}), and logic graphs with $B=1$ (Fig.~\ref{fig:logic_length_stops_branch1}).

 

\subsection{Additional Analysis}\label{app:additional_analysis}

In this section, we provide some additional analysis results on model responses, namely the next step accuracy and completion tokens used. Similar to the error analysis, We categorize these into symbolic, logic graphs with $B>1$, as well as symbolic, logic graphs with $B=1$.

\begin{figure}
     \centering
     \includegraphics[width=\linewidth]{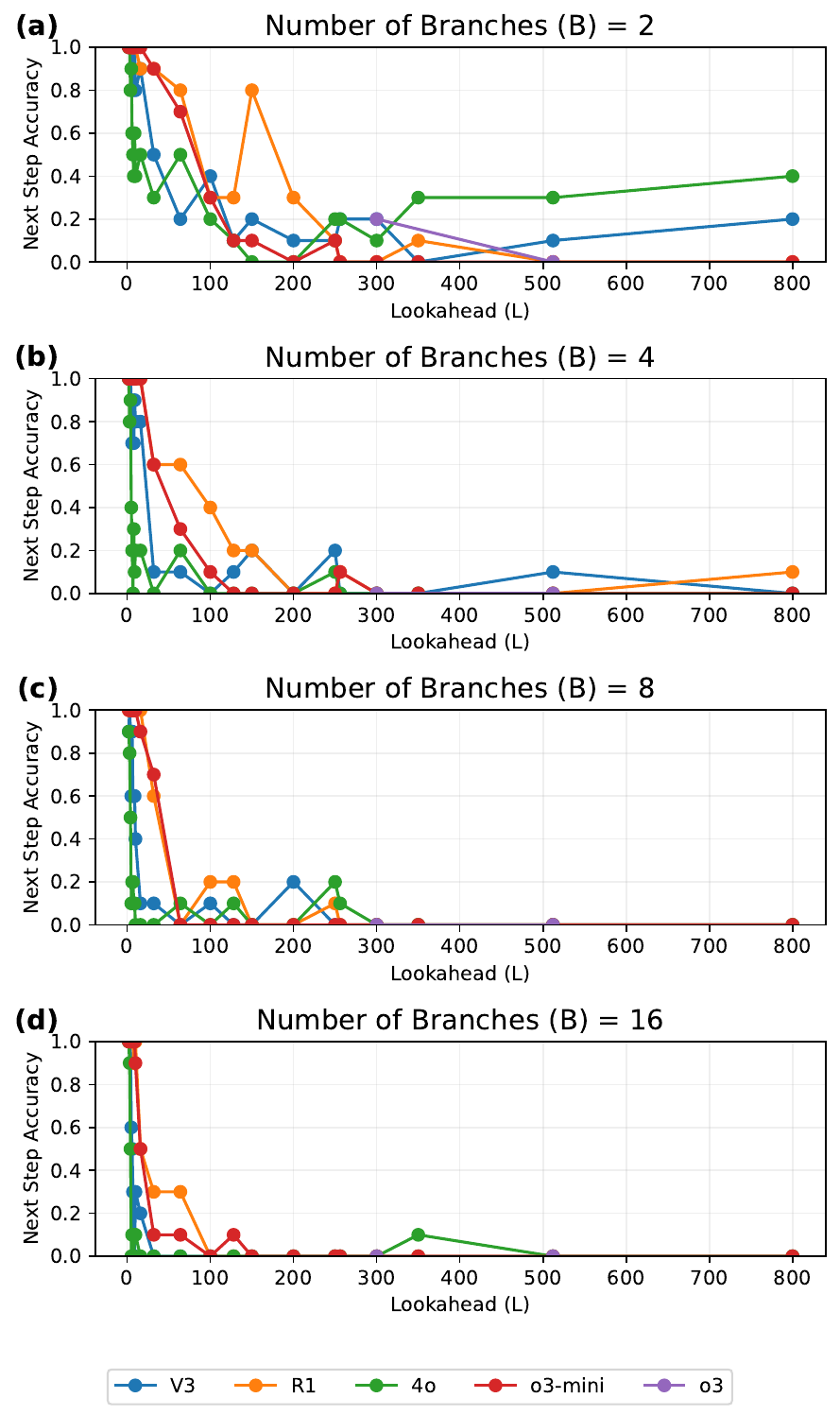}
     \caption{Next step accuracy for symbolic graph with $B > 1$.}
     \label{fig:symbolic_first_step}
 \end{figure}

\begin{figure}
     \centering
     \includegraphics[width=\linewidth]{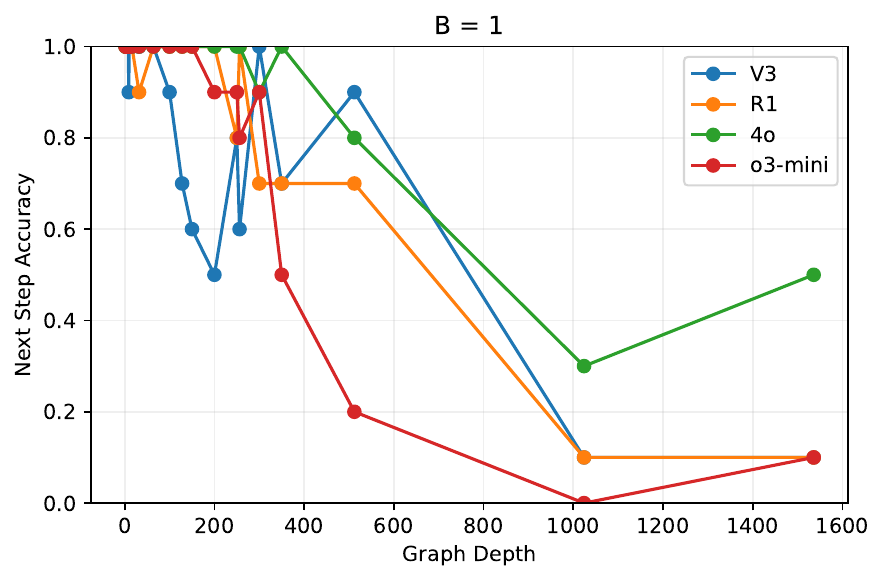}
     \caption{Next step accuracy for symbolic graph with $B = 1$.}
     \label{fig:symbolic_first_step_branch1}
 \end{figure}

\paragraph{Next Step Accuracy}
Next step accuracy indicates the accuracy of the model's prediction when asked to give the next node in the shortest path from start to goal, averaged per lookahead and branch size. For $B>1$ (Fig.~\ref{fig:symbolic_first_step}), accuracy is highest at small lookaheads and generally decreases as lookahead grows. With increasing $B$, the drop in accuracy starts a smaller and smaller and smaller lookaheads.  We observe corresponding behavior for $B=1$ (Fig.~\ref{fig:symbolic_first_step_branch1}).
 
\begin{figure}
     \centering
     \includegraphics[width=\linewidth]{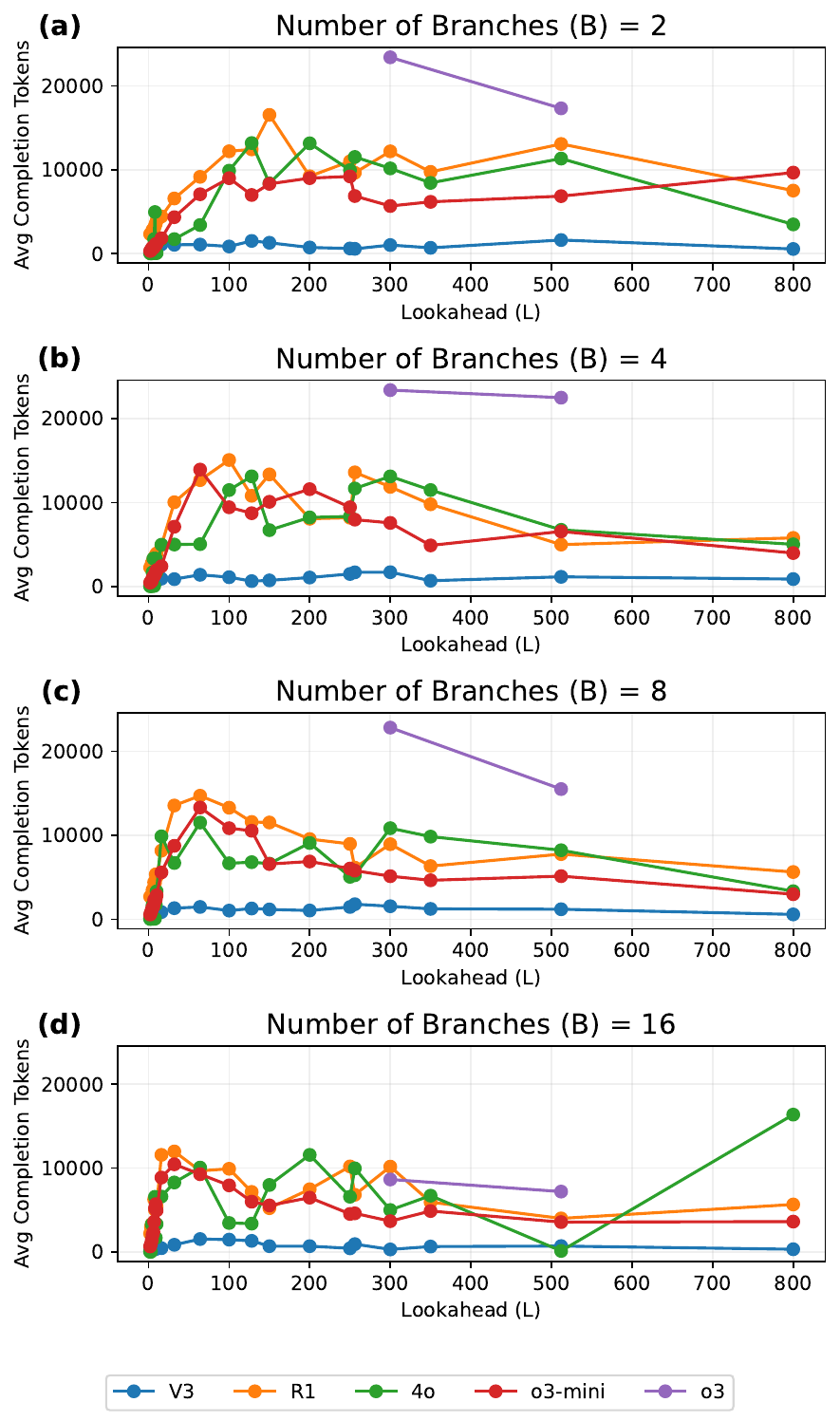}
     \caption{Average completion tokens for symbolic graphs with $B > 1$.}
     \label{fig:symbolic_completion_tokens}
 \end{figure}

\begin{figure}
     \centering
     \includegraphics[width=\linewidth]{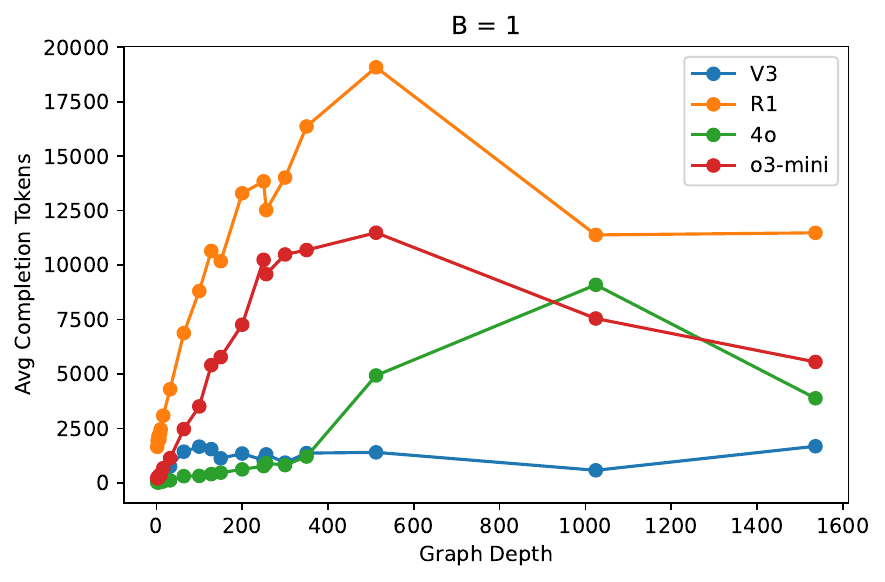}
     \caption{Average completion tokens for symbolic graphs with $B = 1$.}
     \label{fig:symbolic_completion_tokens_branch1}
 \end{figure}

\begin{figure}
     \centering
     \includegraphics[width=\linewidth]{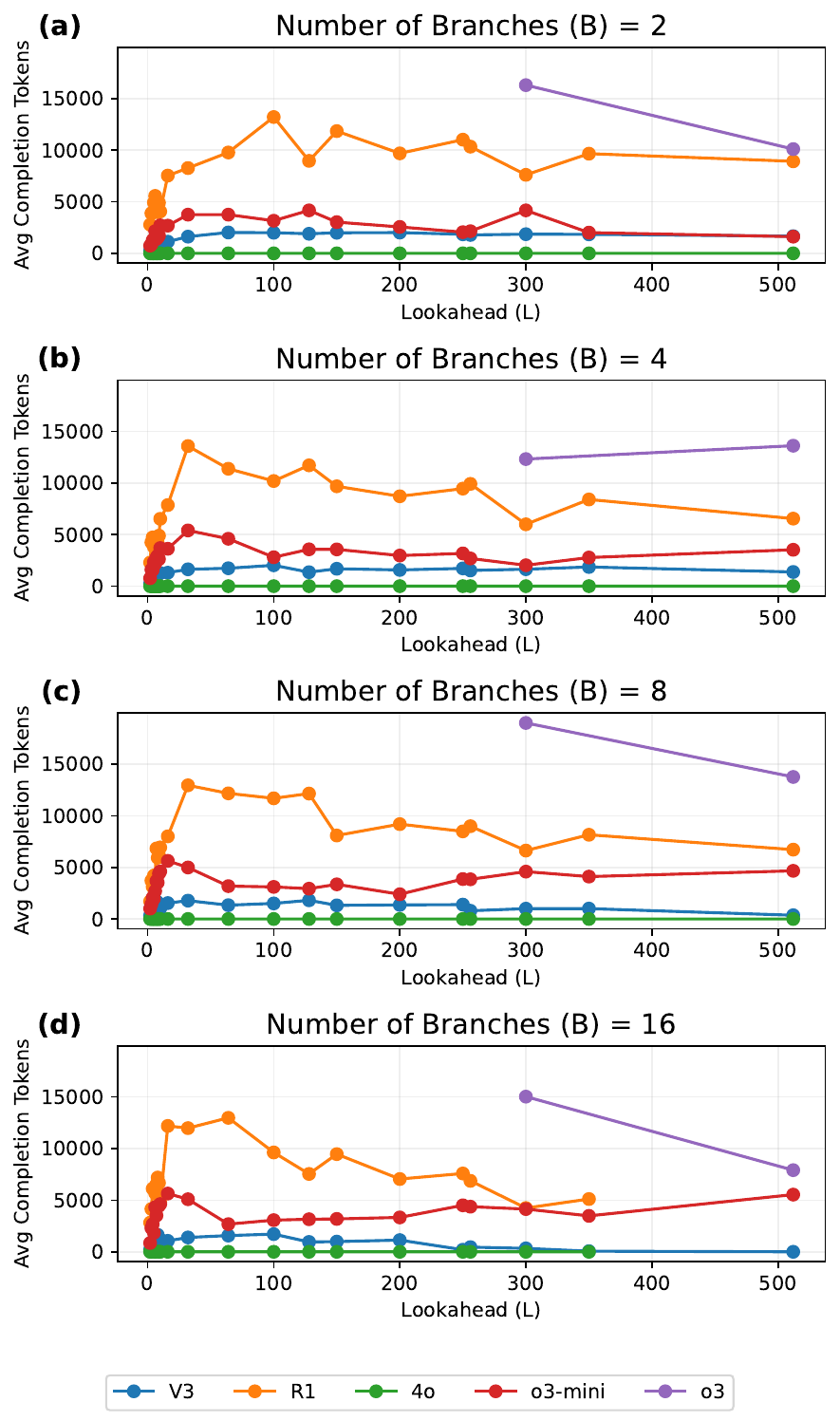}
     \caption{Average completion tokens for logical graphs with $B > 1$.}
     \label{fig:logic_completion_tokens}
 \end{figure}

\begin{figure}
     \centering
     \includegraphics[width=\linewidth]{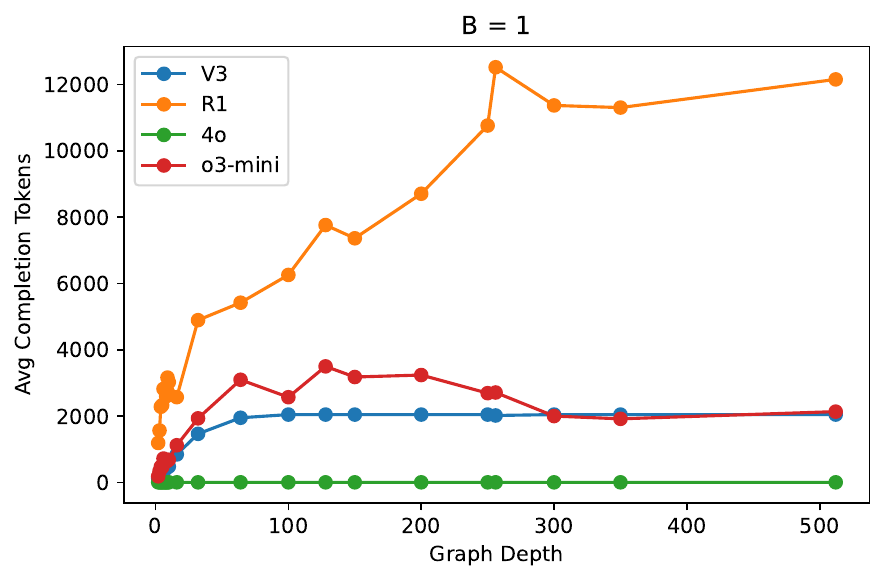}
     \caption{Average completion tokens for logical graphs with $B = 1$.}
     \label{fig:logic_completion_tokens_branch1}
 \end{figure}

\paragraph{Average Completion Tokens}
We plot the average number of tokens generated per prompt, which includes thinking and output tokens for reasoning models, per lookahead and branch size. The amount of tokens used vary drastically across the different models, however, across all models, counts typically rise from small lookahead into a mid-lookahead peak and then taper at larger lookahead. This behavior is consistently observed for symbolic graphs with $B>1$ (Fig.~\ref{fig:symbolic_completion_tokens}), symbolic graphs with $B=1$ (Fig.~\ref{fig:symbolic_completion_tokens_branch1}), logic graphs with $B>1$ (Fig.~\ref{fig:logic_completion_tokens}), and logic graphs with $B=1$ (Fig.~\ref{fig:logic_completion_tokens_branch1}).
 

\subsection{Verification of Real-World Proofs}\label{app:proof_verification}
\begin{figure}
     \centering
     \includegraphics[width=\linewidth]{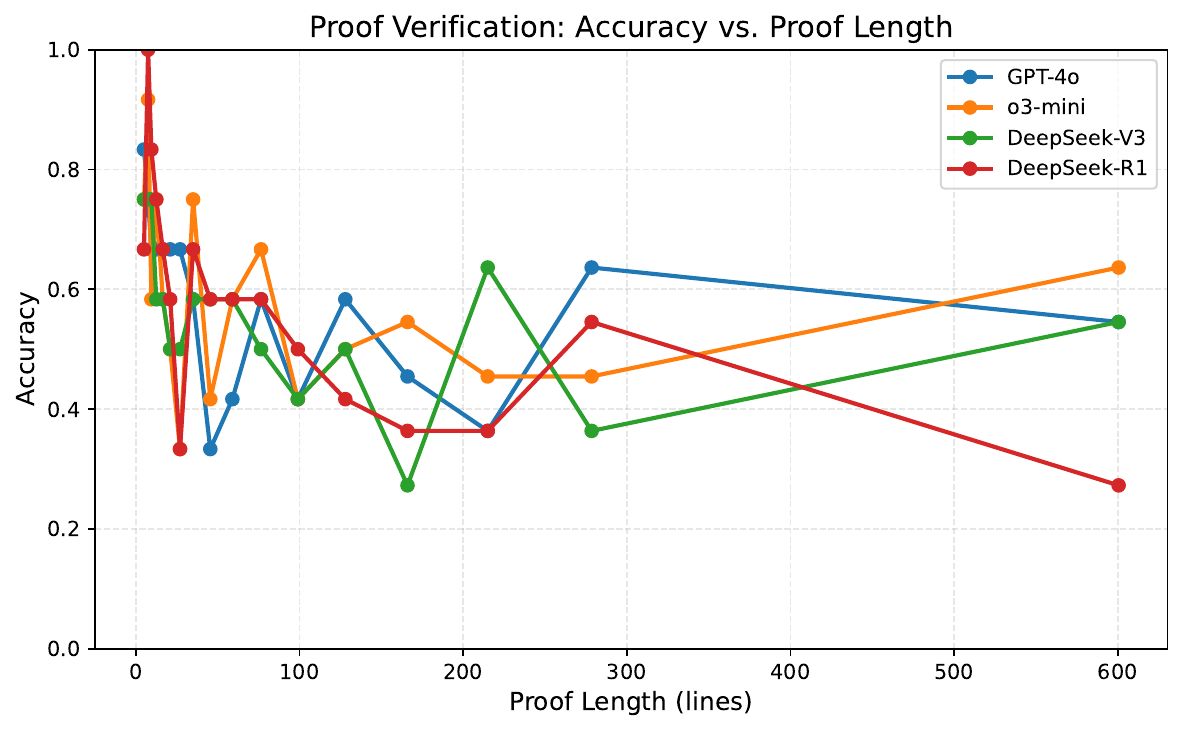}
     \caption{Error detection accuracy of various models, with respect to proof length (lines), when a single perturbation in the proof is introduced. The dataset can be noisy, but we still see a dramatic drop in performance.}
     \label{fig:proof_verfication}
 \end{figure}

\begin{figure}
     \centering
     \includegraphics[width=\linewidth]{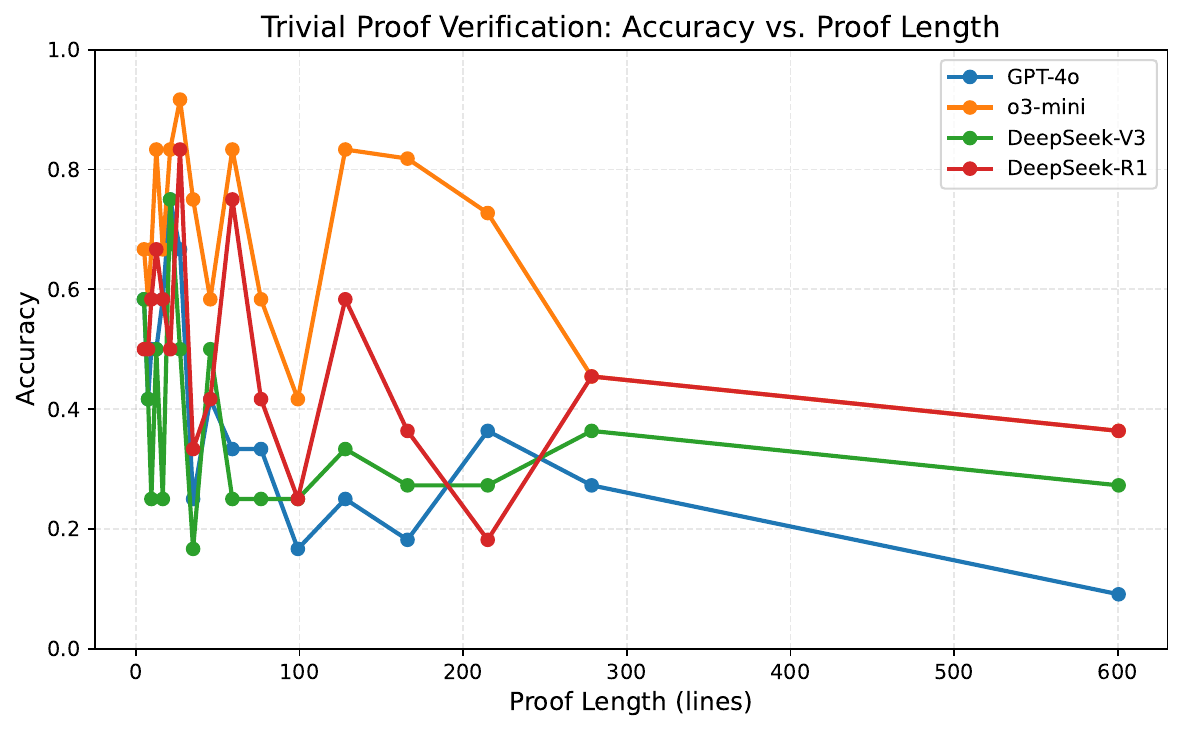}
     \caption{Error detection accuracy of various models, with respect to proof length (lines), when NO perturbation in the proof is introduced. This trivializes the problem, and the expected answer is "no error". We still see an abrupt performance drop.}
     \label{fig:proof_verification_trivial}
 \end{figure}

In this section, we provide an example of perturbations that we introduce in real-world proofs in Table~\ref{tab:proof-perturbation-examples}, along with the error detection accuracy of various models on the trivial proofs (Fig.~\ref{fig:proof_verification_trivial}) and the perturbed proofs (Fig.~\ref{fig:proof_verfication}).

\begin{table}
\centering
\footnotesize 
\renewcommand\arraystretch{1.1}
\begin{tabular}{|>{\ttfamily}p{0.97\linewidth}|}
\hline
Examples of perturbations introduced in proofs\\
\hrulefill\\
Replace ``closed interval'' with ``open interval'' \\

Replace ``$=$'' with ``$\neq$'' \\

Replace ``$n=m$'' with ``$n>m$'' \\

Replace ``Irrational'' with ``Rational'' \\

Replace ``$+$'' with ``$-$'' \\

Replace ``$x^2$'' with ``$x^3$'' \\
... \\ \\
\hline
Prompt: \\
\hline
You are given one proof line. Change the line so that it becomes incorrect but still mathematically sensible.
Do NOT change numbering, labels, formatting-only tokens, or spacing.
Try to keep the line very similar to the original line (so if there are markdown tokens for closing a line or spacing, keep those).
Return ONLY the modified single line (no commentary, no extra lines).
Make only ONE single change for the line. Change only ONE mathematical property or relation, do not do multiple changes. \\ \\
\hline
\end{tabular}
\caption{Examples of perturbations introduced in proofs. We try to make a semantics-altering change to the proof line while still maintaining the surface form of the proof line. They require proof context to detect (all are valid symbols and valid syntax), subtle, and diverse. We also provide the prompt we use to make this perturbation. In future work, this method of perturbation can be improved or parameterized for further analysis.}
\label{tab:proof-perturbation-examples}
\end{table}


\end{document}